\documentclass[a4paper,fleqn]{cas-dc}
\usepackage{dblfnote}
\usepackage[]{natbib}
\usepackage{xcolor}
\usepackage{hyperref}
\usepackage{algorithm}
\usepackage{algorithmic}
\usepackage{multicol}
\usepackage{subfig,graphicx}
\usepackage[T1]{fontenc}
\usepackage{mathdesign}
\usepackage{lipsum}
\usepackage{titlesec}
\usepackage{lscape}
\usepackage{mathrsfs}
\usepackage{booktabs}
\usepackage{multirow}
\usepackage{amsmath}
\usepackage {amssymb}
\usepackage{bbding}
\usepackage{float}
\usepackage[export]{adjustbox}
\usepackage{framed}
\newcommand{\charternumbers}{\fontfamily{bch}\selectfont}
\DeclareTextFontCommand{\textcharter}{\charternumbers}

\titleformat{\section}{\normalfont\fontsize{9pt}{12pt}\bfseries}{\textcharter{\thesection.}}{0.5em}{}
\titleformat{\subsection}{\normalfont\fontsize{9pt}{12pt}\normalfont}{\textcharter{\thesubsection.}}{0.3em}{}
\titleformat{\subsubsection}{\normalfont\fontsize{9pt}{12pt}\normalfont}{\textcharter{\thesubsubsection.}}{0.4em}{}

\titleformat{name=\section,numberless}{\normalfont\fontsize{9pt}{12pt}\bfseries}{}{0em}{}
\titleformat{name=\subsection,numberless}{\normalfont\fontsize{9pt}{12pt}\bfseries}{}{0em}{}
\titleformat{name=\subsubsection,numberless}{\normalfont\fontsize{9pt}{12pt}\bfseries}{}{0em}{}

\usepackage{hyperref}
\newcommand{\doi}[1]{\href{http://dx.doi.org/#1}{\fontsize{8pt}{2pt}\textcharter{\selectfont{#1}}}}
\usepackage{balance}
\usepackage{color}
\hypersetup{
  colorlinks=true,
  linkcolor=cyan,
  urlcolor=cyan,
  citecolor=cyan,
  linktoc=all
}
\usepackage{fancyhdr}
\begin{document}
\thispagestyle{empty}
\let\WriteBookmarks\relax
\def\floatpagepagefraction{1}
\def\textpagefraction{.001}
\let\printorcid\relax


 \shortauthors{BX Jiang et~al.}

\title[mode = title]{Heuristic-enhanced Candidates Selection strategy for GPTs tackle Few-Shot Aspect-Based Sentiment Analysis}  



\author[a]{\textcolor{black}{Baoxing Jiang}}
\ead{2024323045025@stu.scu.edu.cn}
\author[a]{\textcolor{black}{Yujie Wan}}
\ead{2023323045036@stu.scu.edu.cn}
\author[a]{\textcolor{black}{Shenggen Ju}}
\cormark[1]
\ead{jsg@scu.edu.cn} 

\affiliation[a]{organization={College of Computer Science}, 
	addressline={Sichuan University}, 
	city={Chengdu},
	postcode={610005}, 
	country={China}}        
\cortext[1]{Corresponding author} 

\begin{abstract}
Few-Shot Aspect-Based Sentiment Analysis (FSABSA) is an indispensable and highly challenging task in natural language processing. However, methods based on Pre-trained Language Models (PLMs) struggle to accommodate multiple sub-tasks, and methods based on Generative Pre-trained Transformers (GPTs) perform poorly. To address the above issues, the paper designs a Heuristic-enhanced Candidates Selection (HCS) strategy and further proposes All in One (AiO) model based on it. The model works in a two-stage, which simultaneously accommodates the accuracy of PLMs and the generalization capability of GPTs. Specifically, in the first stage, a backbone model based on PLMs generates rough heuristic candidates for the input sentence. In the second stage, AiO leverages LLMs' contextual learning capabilities to generate precise predictions. The study conducted comprehensive comparative and ablation experiments on five benchmark datasets. The experimental results demonstrate that the proposed model can better adapt to multiple sub-tasks, and also outperforms the methods that directly utilize GPTs.
\end{abstract}

\maketitle

\thispagestyle{empty}

\section{Introduction}

\begin{figure}[h]
	\centering
		\scalebox{.95}{
			\includegraphics{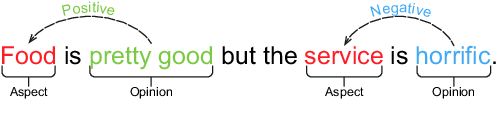}
		}
		\caption{Example sentence with three entities where aspects, opinions, and sentiments are marked in different colors.}
		\label{example}
	\end{figure}

Few-shot Aspect-Based Sentiment Analysis (FSABSA) is an extension of Aspect-Based Sentiment Analysis (ABSA) from a low-resource perspective, and it is rarely addressed due to its lack of data and high processing difficulty \citep{absa_beign}. As shown in Fig. \ref{example}, FSABSA identifies 3 key entities in a sentence \citep{jet,span-aste}: the aspect entity (the subject of sentiment), the opinion entity (the medium of sentiment expression), and the sentiment polarity (the emotional attitude toward the aspect entity).

	FSABSA sub-tasks are mainly divided into multiple types: extraction, classification, and matching. The Aspect Opinion Pair Extraction (AOPE) sub-task \citep{AOPE_1,AOPE_2} focuses on extracting aspect and opinion pairs. The Aspect Level Sentiment Classification (ALSC) sub-task \citep{ALSC_1,ALSC_2} identifies sentiment attitudes toward a given aspect entity. The Aspect Sentiment Triplet Extraction (ASTE) sub-task \citep{peng-two-stage,ASTE_1} extracts and matches aspect and opinion entities. Existing work primarily focuses on designing targeted supervised learning models for a single sub-task, which has achieved significant results. However, these efforts cannot adapt to situations with data scarcity, nor can the models transfer between different tasks. Traditional methods based on pre-trained language models (PLMs) often require specific network structures for related sub-tasks, and data scarcity leads to extremely poor model training results \citep{jbxfewshot,fewshotTE}, which further hinders the development of FSABSA.
	
	It is encouraging that the development of GPTs \citep{GPT3,GPT4,ERNIE} offers a promising solution to FSABSA challenges. However, practical work \citep{harnessing} has shown that directly using GPTs for FSABSA yields minimal results, far below those of solutions designed based on PLMs. This is because GPTs cannot adequately understand sentiment analysis tasks. To fully explore the potential of GPTs in sentiment analysis and simultaneously leverage the excellent performance of PLMs, we designe a Heuristic-enhanced Candidates Selection (HCS) strategy and further develope an All in One (AiO) model. First, a backbone model based on PLMs generates rough candidates for the input sentences, and then carefully designed prompt templates guide the GPTs to make precise predictions. Additionally, we established a rigorous learning paradigm for using GPTs to handle FSABSA, filling a gap in this field. We conducted extensive experiments on 5 public datasets to validate the effectiveness of the proposed model. Compared to using GPTs alone, employing the HCS strategy significantly improved the accuracy of FSABSA processing.
	
	The main contributions in this paper can be summarized as follows:
	\vspace{-3pt}
	\begin{enumerate}
		\item This is the first work to systematically study the use of GPTs for FSABSA. We established a reproducible learning paradigm for GPTs to effectively handle FSABSA, filling a gap in this research field.
		
		\item We propose Heuristic-enhanced Candidates (HCS) Selection strategy and further propose the All in One (AiO) model, and perform rich experiments on 5 FSABSA datasets to illustrate its robust generalization capability and exceptional performance.
		
		\item We present comprehensive research on GPTs applied to FSABSA by implementing ChatGPT-3.5, ERNIE-3.5, and open-source GPT-J. This work provides subsequent researchers with a wealth of reliable experimental results and conclusions. The code is available at \url{https://github.com/BaoSir529/gpt4absa}.
	\end{enumerate}
	
	\section{Related Work}
	
	Over the past few decades, sentiment analysis has been a key area in NLP \citep{absasurvey}, evolving with AI advancements. Initially, it focused on document-level sentiment analysis, resulting in coarse-grained insights \citep{cucaoabsa1,cucaoabsa2}. Deep learning models like CNN and LSTM enabled fine-grained sentiment analysis by capturing sentence-level semantics \citep{absa_beign}. This led to specific tasks such as Aspect Extraction (AE) \citep{ae1,ae2} and Aspect-Level Sentiment Classification (ALSC) \citep{jbxalsc}.
	
	Most of the existing approaches are geared towards aspect based sentiment analysis (ABSA) in a full-resource data perspective, rather than FSABSA. The rise of pre-trained language models (PLMs) like BERT \citep{transformer,bert} revolutionized NLP, making fine-tuning the dominant ABSA approach. PLMs enabled joint tasks due to their robust semantic capabilities. For instance, AOPE \citep{aoe1,aoe2,aoe3} matched aspect-opinion pairs, while ALSC \citep{alsc2,alsc3} determined sentiment from specific aspects. Recent breakthroughs with GPT models \citep{GPT3,ERNIE} have significantly impacted NLP methodologies. GPT-based prompt-tuning has been applied successfully in various research areas \citep{kbvqa1,kbvqa2}. GPT's effectiveness in few-shot learning has spurred interest in ABSA. \citet{chatagri} used GPT-3.5 for few-shot training in agricultural text classification. \citet{ITMIT} addressed FSABSA sub-tasks using instruction-based tuning, and \citet{NAPT} explored weakly supervised learning for FSABSA based on the T5 model. \citet{202401} attempted to solve FSABSA with in-context learning paradigm.
	
	Despite these advancements, FSABSA lacked a comprehensive model for multiple sub-tasks and multi-domain datasets. This paper defines a general GPT-based prompt-tuning paradigm for ABSA and introduces the AiO model, which efficiently completes FSABSA sub-tasks (AOPE, ALSC, ASTE).
	
	\section{Approach}
	
	\subsection{GPT In-context Learning}
	The GPT model has proven to be a powerful natural language processing tool, boasting impressive in-context few-shot learning capabilities. Unlike traditional techniques that require fine-tuning a PLMs for specific tasks, the GPT paradigm's in-context few-shot learners quickly adjust to new tasks during inference with minimal examples, thereby eliminating the need for parameter updates.
	
	Specifically, the model predicts the target of the new task based on the fixed prompt head $\mathcal{H}$, given context $\boldsymbol{C} = \{c_1, c_2, \dots, c_n\}$ and the new task's input $x$, treating it as a text sequence generation task. The target $\boldsymbol{y}=(y^1, y^2, \dots, y^L)$ is computed by the following equation:
	\begin{equation}
		y^t = \underset{\hat{y}^t}{{argmax}}\; p(\hat{y}^t|\mathcal{H},\boldsymbol{C},x,y^{<t}),
	\end{equation}
	where each $y^t$ is the argmax of the probability distribution predicted by the language model conditioned on context $\boldsymbol{C}$, input $x$, and previous predictions $y^{<t}$.
	
	\subsection{GPTs for FSABSA}
	
	Given a sentence $\boldsymbol{S}=(t_1, t_2, \dots, t_n)$, which categorizes entities into aspect entities $\boldsymbol{A}=\{(t_i, \dots, t_j)|0\leqslant i\leqslant j \leqslant n\}$ and opinion entities $\boldsymbol{O}=\{(t_i, \dots, t_j)|0 \leqslant i \leqslant j \leqslant n\}$. Classify sentiment polarity $\boldsymbol{P}=\{ {Negative},  \\ {Neutral},  {Positive}\}$ according to the sentence. Specifically, the definitions of each FSABSA sub-task are outlined as follows:
	\begin{itemize}
		\item{Aspect-Level Sentiment Classification (ALSC) entails discerning sentiment polarity $p_i$ expressed in the sentence regarding the given aspect entities. \textbf{Input:} $(\boldsymbol{S}, \boldsymbol{a})$, \textbf{Output:} $\{\boldsymbol{p}|p_i \in \boldsymbol{P}\}$.}
		
		\item {Aspect-Opinion Pair Extraction (AOPE) identify all pairs that consist of an aspect term $a_i$ and corresponding opinion term $o_i$. \textbf{Input:} $\boldsymbol{S}$, \textbf{Output:} $\{(a_i, o_i) | a_i \in \boldsymbol{A}, o_i \in \boldsymbol{O}\}$.}
		
		\item{Aspect Sentiment Triplet Extraction (ASTE) needs to extract aspect entities $a_i$ and opinion entities $o_i$ from sentences while determining the corresponding sentiment polarity $p_i$. \textbf{Input:} $\boldsymbol{S}$, \textbf{Output:} $\{(a_i, o_i, p_i) | a_i \in \boldsymbol{A}, o_i \in \boldsymbol{O}, p_i \in \boldsymbol{P}\}$.}
	\end{itemize}
	The primary challenge in applying GPTs to FSABSA lies in enhancing the model’s accuracy and comprehensiveness in sentiment analysis while harnessing its robust context-learning capabilities. With this objective in mind, we further define a novel general learning paradigm for GPTs tackle FSABSA.
	\begin{enumerate}
		\item{\textbf{Prompt Template Construction:} Suitable prompt templates are constructed to assist GPT in quickly adapting to downstream sub-task content. This ensures the generation of diverse responses that align with specific sub-task requirements.}
		
		\item{\textbf{Heuristic-enhanced Candidates Selection:} PLM-based models designed for FSABSA determine a set of heuristic-enhanced candidates for GPTs. This selection aims to maximally aid GPT in adapting to downstream sub-task content.}
		
		\item{\textbf{GPT Inference:} The inference process of GPT is executed to allow the model to autonomously generate text responses for downstream sub-tasks, typically requiring minimal intervention.}
		
		\item{\textbf{Answer Normalization or Alignment:} The model-generated responses are standardized or aligned to ensure they map to the expected answer format for specific sub-tasks. This facilitates evaluating the model’s performance.}
	\end{enumerate}
	
	\subsection{The AiO Framework}
\begin{figure*}[t]
		\centering
		\scalebox{1.6}{
			\includegraphics{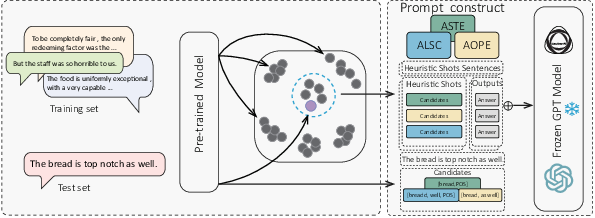}
		}
		\caption{Overview of AiO model. The left part is heuristic-enhanced candidates selection stage and the right part is answer inference stage.}
		\label{model}
	\end{figure*}

	The AiO framework we propose is a conceptually simple yet efficient two-stage model. A model based on PLMs designed for ABSA generates heuristic candidates in stage one. Subsequently, appropriate shots are chosen for the test input based on specific filtering rules. In the stage two, the task-specific prompt head, filtered candidate shots, corresponding heuristic responses, and test input are integrated into a formatted prompt template, guiding the GPTs to generate the correct answer. An overview of the AiO structure is shown in Fig. \ref{model}.
	
	\subsubsection{Stage 1. Heuristic-Enhanced Candidates Selection}
	
	The heuristic-enhanced candidates selection (HCS) strategy aims to derive a set of heuristic answers and a series of heuristic shots for the AiO samples input through a PLMs-based ABSA model. In essence, most ABSA models $\mathcal{M}$ based on PLMs can bifurcate into an encoding network $\mathcal{M}_e$ tasked with acquiring semantic information from the input text and a trainable decoding network $\mathcal{M}_c$ designed to adapt to downstream tasks.
	
	Assuming all sub-tasks constitute a set $\varsigma = [{AOPE}, {ALSC},\\ {ASTE}]$, define any FSABSA dataset as $\mathcal{D}=\{\boldsymbol{D}^i\}_{i=1}^M$, the selection must satisfy the following criteria:
	\begin{align}
		&\forall \boldsymbol{D}^i=( \boldsymbol{S}^i, \boldsymbol{T}^i)\\
		&\forall \boldsymbol{T}^i=(a_i,o_i,p_i) \in \vec{\mathbb{I}}(\varsigma_i) \cdot (\boldsymbol{A}_i, \boldsymbol{O}_i, \boldsymbol{P}_i)\\
		&\vec{\mathbb{I}}(\varsigma_i)[j]=\left\{ \begin{array}{l}
			1, {Input}_{\varsigma}[j] \ne None\\
			0, {otherwise};\\
		\end{array} \right. ,
	\end{align}
	where $\boldsymbol{S}^i$ indicate the sentence sets, $\boldsymbol{T}^i$ refers to the sentiment triplet $(a_i,o_i,p_i)$ in each sentence, $\boldsymbol{A},\boldsymbol{O},\boldsymbol{P}$ represent the sets of aspect entities, opinion entities, and sentiment polarities as defined earlier. $\varsigma$ denotes the specific task type, $ {Input}_{\varsigma}[j]$ denotes the $j$-th input for the corresponding task, and $\vec{\mathbb{I}}(\cdot)$ represents the indicator function only taking value $1$ when the condition is satisfied.
	
	Within the dataset, encoding network $\mathcal{M}_e$ extracts the semantic feature of the training data. The acquisition of semantic features $\boldsymbol{Z}^i$ can be succinctly denoted by the following process:
	\begin{equation}
		\boldsymbol{Z}^i = \mathcal{M}_e(\boldsymbol{S}^i),
	\end{equation}
	
	The decoding network $\mathcal{M}_c$ is typically designed to adapt to downstream sub-tasks. The decoder consists of a sequence of Feed-Forward Neural Networks (FFNN) and activation functions (e.g., SoftMax). The decoding process of sub-tasks with respect to semantics can be succinctly denoted as:
	\begin{equation}
		\boldsymbol{R}^i = \mathcal{M}_c (\boldsymbol{Z}^i).
	\end{equation}
	where $\boldsymbol{R}^i=(\hat{a}_i,\hat{o}_i,\hat{p}_i)$, it needs to satisfy the conditions for $\boldsymbol{T}^i$ to hold.
	
	In practice, the results generated by model $\mathcal{M}$ only hold practical significance when the training converges or the loss value reaches a predefined threshold. Thus, we first completely trained an independent model and aggregated its results on the training set into a list $\mathcal{R}=\{\boldsymbol{R}^i\}_{i=1}^N$. Note that $N<M$, indicating that $\mathcal{R}$ contains model outputs only for the training samples. After obtaining the initial predictions for the dataset, we utilize the trained model and these results to generate appropriate heuristic answers and heuristic shots and construct prompt templates for input samples for AiO. We describe the generation of the two candidates in detail below.
	
	\textbf{Heuristic Answers} first establish a series of possible predictions, providing constraints for the inference model’s response. For any given sub-task, the model may generate multiple rough predictions, which can help the GPTs make more accurate inferences. In prior work \citep{kbvqa1}, this has been demonstrated to yield reliable effects. For a specific sample $\boldsymbol{D}_p \in \mathcal{D}$, we can feed it into a well-trained model $\mathcal{M}$ to obtain the predictions based on the given sub-task $\varsigma_i$ , and then chooses the meaningful results $ \boldsymbol{R}_p $ as the heuristic answers.
	\begin{equation}
		\boldsymbol{R}_p = \mathcal{M}(\boldsymbol{D}_p, \varsigma_i).
	\end{equation}
	
	\textbf{Heuristic Shots} aim to select the best contextual examples from the training set for each instance of inference based on the effectiveness of few-shot prompt learning in helping the model adapt quickly to downstream tasks \citep{kbvqa1,kbvqa2}. An appropriate context shot $\boldsymbol{D}_s$ should have a similar expression and object of expression as the input sample $\boldsymbol{D}_p$. Specifically, we construct the encoder $E$ composed of BERT to semantically map the text information from the input sample $\boldsymbol{S}_p \in \boldsymbol{D}_p$ and all contextual examples $\boldsymbol{S}_s^i \in \boldsymbol{D}_s^i$ to an entity space, yielding text embeddings $\boldsymbol{e}_p$ and $\boldsymbol{e}_s^i$:
	\begin{align}
		\boldsymbol{e}_p &= E(\boldsymbol{S}_p)\\
		\boldsymbol{e}_s^i &= E(\boldsymbol{S}_s^i).
	\end{align}
	
	After that, we calculated the cosine similarity between $\boldsymbol{D}_p$ and all $\boldsymbol{D}_s$ to select the top $K$ examples with the highest similarity as heuristic shots for the current sample:
	\begin{equation}
		\mathcal{I}_p=\underset{i\in \{1,\dots,K\}}{argTopK}\frac{\boldsymbol{e}_p \cdot \boldsymbol{e}_s^i}{\Vert \boldsymbol{e}_p \Vert \Vert \boldsymbol{e}_s^i \Vert},
	\end{equation}
	where $\mathcal{I}_p$ is the index set of the top $K$ samples most similar to the input sample within the training set $\{\boldsymbol{D}_{i}\}_{i=1}^K \subseteq \mathcal{D}$. 
	
	Subsequently, the heuristic shots template for $\boldsymbol{D}_p$ were constructed as follows:
	\begin{equation}
		\mathcal{T}(\boldsymbol{D}_p) = \{\boldsymbol{D}^i,\boldsymbol{R}^i | i \in \mathcal{I}_p\}.
	\end{equation}
	where $\boldsymbol{R}^i$ refers to the heuritic answers of the $i$-th shot as described earlier.
	
\subsubsection{Stage 2. Answer Inference}
	In this stage, prompt templates are created for the input samples specific to the various sub-tasks, facilitating a quick response from the GPT. For a given input, AiO selects a specific prompt head template, $h_i$, based on sub-task type $\varsigma_i$, and then creates a heuristic prompt by combining the heuristic answer and heuristic shots obtained in stage 1.

\begin{figure}
	\centering
		\resizebox{\columnwidth}{!}{\includegraphics{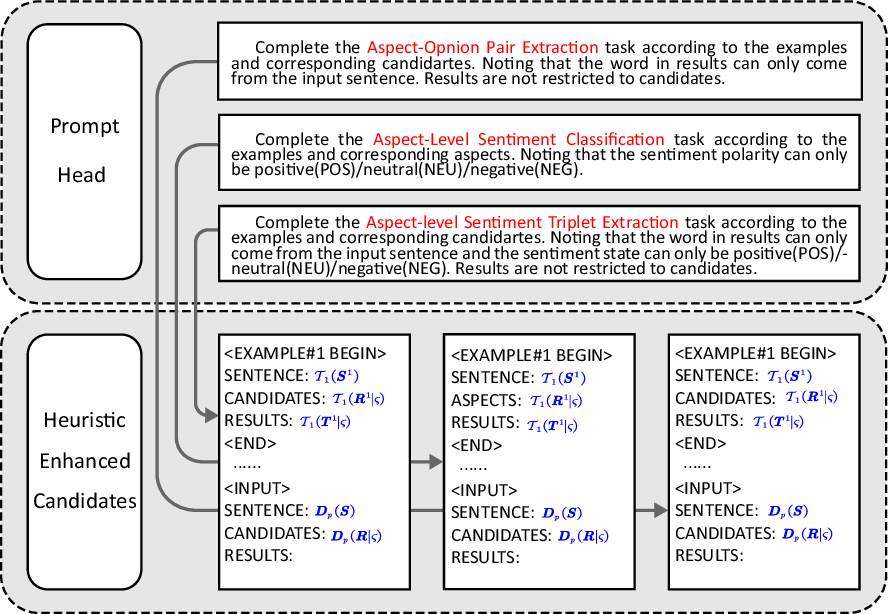}}
		\caption{Example of constructing a prompt template based on a particular sub-task $\varsigma$.}
		\label{prompt}
	\end{figure}

	A well-crafted prompt head template has been proven to swiftly guide the reasoning capabilities of the GPT model toward a particular aspect \citep{kbvqa1,chatagri}. Through research into numerous pioneering works \citep{now1,now2}, we manually modified the prompt words generated based on promptperfect\footnote{\url{https://promptperfect.jina.ai/}} to create templates. To enhance the readability and comprehension of the prompt template by the model, we incorporated certain fixed-format text, such as "EXAMPLE \#1". For instance, given an input sample $\boldsymbol{D}_p$ and its specific sub-task $\varsigma_i$, the construction of the prompt sample is depicted in Fig. \ref{prompt}.
	To fully utilize the available examples, we followed the experience of \citep{kbvqa1} and introduced a multi-query ensemble strategy. Specifically, for a given input, $T \times K$ heuristic shots are selected to form $T$ sets of prompt templates. After parallel inference, the model obtains $T$ sets of distinct predictions. Since each inference group may yield multiple diverse results, AiO selects the union as the final prediction to preserve result diversity.

	\section{Experiment Details}
	\subsection{Hyperparameter Settings}
	We used the dualencoder4aste \citep{jbxaste} as a backbone for fine-tuning each dataset. To ensure fairness in our experiments and to reduce variability in the results due to variations in the generalization of the backbone model, we set the same random seed for all datasets involved in the evaluation. In addition, we ran the backbone model with the same hyperparameters 10 times for each dataset and selected the model with mid-range performance as the final backbone. In addition, to provide more comprehensive coverage of the experimental results, we used several popular GPT models, including ChatGPT-3.5, ERNIE-3.5, and GPT-J. This selection covers a range of mainstream open-source and proprietary models.
	
	\subsection{Datasets}
	We validated the proposed AiO model on 5 datasets. SemEval V2 dataset \citep{jet} is the most popular and widely used in the ABSA field. This dataset comprises 4 sub-datasets: L\small AP\normalsize14 \citep{semeval-2014-task4}, R\small EST\normalsize14 \citep{semeval-2014-task4}, R\small EST\normalsize15 \citep{semeval-2015-task12}, and R\small EST\normalsize16 \citep{semeval-2016-task5}. The initial data for these sub-datasets originated from specific tasks in past SemEval challenges and were subsequently refined and organized by researchers \citep{fan-first-label-dataset,peng-two-stage} to create a stable version. This dataset contains predominantly review text related to the laptop and restaurant domains with aspect entities, opinion entities, and sentiment polarities labeled for each review.
	
	MAMS dataset \citep{MAMS} is another challenging FSABSA dataset from the Citysearch New York dataset \citep{MAMS_cite}. Each sentence in this dataset contains an average of 2.62 aspect terms. Every sentence includes opposing sentiment labels, making it particularly challenging to determine the sentiment attitudes toward different aspect entities. Since we are not discussing aspect labels in this paper, we utilized only the ATSA version set from MAMS.
	
	The statistical information for all datasets is provided in Table \ref{statistic}. $\#S$ and $\#A$ indicate the number of sentences and aspects respectively.
	\begin{table}[h]
		\centering
		\caption{Statistics of the datasets. $\#S$ and $\#A$ indicate the number of sentences and aspects, respectively.}
			\scalebox{0.7}{
				\begin{tabular}{@{}lllllllllll@{}}
					\toprule
					\multirow{2}{*}{Dataset} & \multicolumn{2}{l}{LAP14} & \multicolumn{2}{l}{RES14} & \multicolumn{2}{l}{REST15} & \multicolumn{2}{l}{RES16} & \multicolumn{2}{l}{MAMS} \\ \cmidrule(l){2-11} 
					& \#S & \#A & \#S & \#A & \#S & \#A & \#S & \#A & \#S & \#A \\ \midrule
					Train & 906 & 1218 & 1266 & 2051 & 605 & 862 & 857 & 1198 & 4297 & 11186 \\
					Dev & 219 & 295 & 310 & 500 & 148 & 213 & 210 & 296 & 500 & 1332 \\
					Test & 328 & 463 & 4992 & 848 & 322 & 432 & 326 & 452 & 500 & 1336 \\ \bottomrule
				\end{tabular}%
			}
			\label{statistic}
		\end{table}
			
\subsection{Experiment Results}
We chose the $F1$ score as the evaluation metric for all sub-tasks. Precision ($P$) and recall ($R$) are also provided for the experimental results. The relationship between these metrics is as follows:
\begin{equation}
F1 = \frac{2\times P\times R}{P + R}.
\end{equation}

	For ALSC sub-task, as shown in Table \ref{alsc}. In \textit{LAP14} dataset, all models showed improvement in $F1$ scores with an increasing number of shots. GPT-J consistently outperformed GPT and ERNIE-3.5 by a substantial margin across all shot levels, with an $F1$ score of 93.27\% at 15 shots. ERNIE-3.5 may have lagged because it was trained primarily on the Chinese dataset, while ChatGPT-3.5 showed a commendable performance increase with additional shots. In \textit{RESTAURANT} datasets, GPT-J maintained a performance edge, indicating robust generalization. ERNIE-3.5 and ChatGPT-3.5 were closely matched, with ChatGPT-3.5 improving more with increased shots. In \textit{MAMS} dataset, GPT-J was still the highest, showing adaptability with limited data. ERNIE-3.5 and ChatGPT-3.5 were closely matched. This emphasizes the difficulty of this dataset.

	\begin{table}[h]
	\centering
	\renewcommand{\arraystretch}{1}
	\setlength{\abovecaptionskip}{0pt}
	\setlength{\belowcaptionskip}{10pt}
	\begin{minipage}{1\linewidth}
		\caption{Experimental results for the ALSC sub-task. Optimal values in each dataset are in bold, and sub-optimal values are underlined.}
		\label{alsc}
		\resizebox{\columnwidth}{!}{
			\scalebox{1}{
				\begin{tabular}{ccccccccccc}
					\hline
					\multirow{2}{*}{Dataset} & \multirow{2}{*}{SHOTS} & \multicolumn{3}{c}{ChatGPT-3.5} & \multicolumn{3}{c}{ERNIE-3.5} & \multicolumn{3}{c}{GPT-J} \\ \cline{3-11} 
					&  & F1 & P & R & F1 & P & R & F1 & P & R \\ \hline
					\multirow{4}{*}{LAP14} 
					& 1 & 80.82 & 81.09 & 80.56 & 31.75 & 31.12 & 32.40 & 62.60 & 56.77 & 69.76 \\
					& 5 & 83.50 & 83.41 & \underline{83.59} & 34.17 & 33.00 & 35.42 & 72.98 & 68.43 & 78.19 \\
					& 10 & \underline{83.77} & \underline{83.95} & \underline{83.59} & 34.30 & 32.87 & 35.85 & 79.92 & 77.48 & 82.51 \\
					& 15 & \textbf{85.26} & \textbf{85.75} & \textbf{84.77} & 32.96 & 31.50 & 34.56 & 80.17 & 79.16 & 81.21 \\ \hline
					\multirow{4}{*}{RES14}
					& 1 & 88.65 & 88.86 & 88.44 & 38.16 & 36.80 & 39.62 & 68.21 & 64.20 & 72.76 \\
					& 5 & \textbf{90.93} & \textbf{91.42} & \textbf{90.45} & 46.36 & 44.52 & 48.35 & 79.46 & 76.39 & 82.78 \\
					& 10 & \underline{90.18} & \underline{90.50} & \underline{89.86} & 46.24 & 44.41 & 48.23 & 83.51 & 81.18 & 85.97 \\
					& 15 & 90.05 & 90.48 & 89.62 & 46.16 & 44.26 & 48.23 & 84.66 & 83.16 & 86.20 \\ \hline
					\multirow{4}{*}{RES15} 
					& 1 & 85.22 & 85.02 & 85.42 & 31.88 & 30.75 & 33.10 & 68.67 & 62.06 & 76.85 \\
					& 5 & 88.48 & 88.07 & 88.89 & 37.65 & 36.09 & 39.35 & 76.77 & 71.69 & 82.64 \\
					& 10 & \underline{88.71} & \underline{88.30} & \underline{89.12} & 35.80 & 34.25 & 37.50 & 82.18 & 79.18 & 85.42 \\
					& 15 & \textbf{89.76} & \textbf{89.24} & \textbf{90.28} & 35.66 & 34.18 & 37.27 & 81.65 & 78.59 & 84.95 \\ \hline
					\multirow{4}{*}{RES16} 
					& 1 & 90.85 & 90.55 & 91.15 & 32.48 & 31.40 & 33.63 & 74.34 & 68.87 & 80.75 \\
					& 5 & 90.19 & 89.89 & 90.49 & 36.34 & 34.60 & 38.27 & 80.50 & 75.78 & 85.84 \\
					& 10 & \underline{92.86} & \underline{92.16} & \underline{93.58} & 39.58 & 37.75 & 41.59 & 82.96 & 79.51 & 86.73 \\
					& 15 & \textbf{93.27} & \textbf{92.97} & \textbf{93.58} & 36.38 & 34.67 & 38.27 & 83.42 & 80.75 & 86.28 \\ \hline
					\multirow{4}{*}{MAMS} 
					& 1 & 60.52 & 60.86 & 60.18 & 36.14 & 36.13 & 36.15 & 34.39 & 33.03 & 35.85 \\
					& 5 & 62.96 & 63.13 & 62.80 & 36.87 & 36.85 & 36.90 & 43.10 & 42.37 & 43.86 \\
					& 10 & \underline{64.24} & \underline{64.48} & \underline{64.00} & 38.67 & 38.64 & 38.70 & 45.37 & 44.58 & 46.18 \\
					& 15 & \textbf{66.60} & \textbf{66.72} & \textbf{66.47} & 39.12 & 39.09 & 39.15 & 45.48 & 44.72 & 46.27 \\ \hline
				\end{tabular}
			}
		}
	\end{minipage}
\end{table}

	For AOPE subtask, as shown in Table \ref{aope}. In \textit{LAP14} dataset, GPT-J demonstrated superior $F1$ scores across shot levels, notably improving from 1 to 15 shots. ERNIE-3.5 and ChatGPT-3.5 had competitive results, with ERNIE-3.5 slightly outperforming ChatGPT-3.5 at lower shot levels. In \textit{RESTAURANT} datasets, GPT-J consistently outperformed GPT and ERNIE-3.5 across the RES14 and RES15 datasets, showing its ability to handle AOPE tasks. ChatGPT-3.5 was competitive with ERNIE-3.5, suggesting its effectiveness in this sub-task.
	
	\begin{table}[h]
		\centering
		\renewcommand{\arraystretch}{1}
		\setlength{\abovecaptionskip}{0pt}
		\setlength{\belowcaptionskip}{10pt}
		\begin{minipage}{1\linewidth}
			\caption{Experimental results for the AOPE sub-task. Optimal values in each dataset are in bold, and sub-optimal values are underlined.}
			\label{aope}
			\resizebox{\columnwidth}{!}{
				\scalebox{1}{
					\begin{tabular}{ccccccccccc}
						\hline
						\multirow{2}{*}{Dataset} & \multirow{2}{*}{SHOTS} & \multicolumn{3}{c}{ChatGPT-3.5} & \multicolumn{3}{c}{ERNIE-3.5} & \multicolumn{3}{c}{GPT-J} \\ \cline{3-11} 
						&  & F1 & P & R & F1 & P & R & F1 & P & R \\ \hline
						\multirow{4}{*}{LAP14} 
						& 1 & 46.35 & 42.08 & 51.57 & 26.77 & 26.92 & 26.62 & 55.01 & 56.29 & 53.79 \\
						& 5 & 56.8 & 47.30 & 71.16 & 24.79 & 24.81 & 24.77 & 62.41 & \underline{64.50} & 60.44 \\
						& 10 & 57.27 & 47.83 & \underline{71.35} & 24.38 & 24.35 & 24.40 & \underline{62.88} & 64.47 & 61.37 \\
						& 15 & 58.88 & 48.84 & \textbf{74.12} & 13.35 & 23.42 & 23.29 & \textbf{64.35} & \textbf{65.33} & 63.40 \\ \hline
						\multirow{4}{*}{RES14} 
						& 1 & 60.07 & 56.23 & 64.49 & 41.03 & 39.79 & 42.35 & 65.21 & 65.85 & 64.59 \\
						& 5 & 66.14 & 58.81 & 75.55 & 40.18 & 39.07 & 41.35 & 69.77 & 69.42 & 70.12 \\
						& 10 & 66.93 & 58.97 & \underline{77.36} & 39.55 & 38.43 & 40.74 & \underline{71.07} & \textbf{70.72} & 71.43 \\
						& 15 & 67.80 & 59.98 & \textbf{77.97} & 39.80 & 38.64 & 41.05 & \textbf{71.13} & \underline{69.69} & 72.64 \\ \hline
						\multirow{4}{*}{RES15} 
						& 1 & 40.55 & 33.25 & 51.96 & 33.90 & 31.35 & 36.91 & 54.07 & 50.54 & 58.14 \\
						& 5 & 49.21 & 38.81 & 67.22 & 36.26 & 32.89 & 40.41 & 60.24 & \underline{55.14} & 66.39 \\
						& 10 & 54.16 & 42.57 & \underline{74.43} & 33.89 & 30.76 & 37.73 & \underline{60.39} & 55.10 & 66.80 \\
						& 15 & 55.82 & 44.22 & \textbf{75.67} & 34.41 & 31.06 & 38.56 & \textbf{61.72} & \textbf{56.06} & 68.66 \\ \hline
						\multirow{4}{*}{RES16} 
						& 1 & 55.44 & 49.84 & 62.45 & 31.95 & 31.07 & 32.88 & 65.85 & 64.09 & 67.70 \\
						& 5 & 63.30 & 53.81 & 76.85 & 31.45 & 30.47 & 32.49 & 69.31 & 67.34 & 71.40 \\
						& 10 & 65.31 & 56.16 & \underline{78.02} & 30.91 & 29.98 & 31.91 & \underline{69.50} & \textbf{67.52} & 71.60 \\
						& 15 & 66.24 & 56.53 & \textbf{79.96} & 31.81 & 30.63 & 33.07 & \textbf{70.27} & \underline{67.44} & 73.35 \\ \hline
					\end{tabular}
				}
			}
		\end{minipage}
	\end{table}
	For ASTE  sub-task, as shown in Table \ref{aste}. In \textit{LAP14} dataset, GPT-J had excellent $F1$ scores in ASTE. ERNIE-3.5 and ChatGPT-3.5 were matched in performance. In \textit{RESTAURANT} datasets, GPT-J consistently outperformed ERNIE-3.5 and ChatGPT-3.5 in ASTE across the RES14 and RES15 datasets. The evaluation results across various sub-tasks were surprising. Compared to larger models like ChatGPT-3.5 and ERNIE-3.5, GPT-J performed better across datasets and model sizes. Our analysis suggests that supervised learning helps models like GPT-J, which have less pre-existing knowledge, adapt better to small-sample data in downstream tasks. In contrast, models trained on extensive datasets, like ChatGPT-3.5, are less affected by small-sample data. For developing large generative models tailored for FSABSA, our observations suggest that smaller models with downstream adaptability training can yield better results than costly large pre-trained models like GPT.
	
	\begin{table}[h]
		\centering
		\renewcommand{\arraystretch}{1}
		\setlength{\abovecaptionskip}{0pt}
		\setlength{\belowcaptionskip}{10pt}
		\begin{minipage}{1\linewidth}
			\caption{Experimental results for the ASTE sub-task. Optimal values in each dataset are in bold, and sub-optimal values are underlined.}
			\label{aste}
			\resizebox{\columnwidth}{!}{
			\scalebox{1}{
			\begin{tabular}{ccccccccccc}
				\hline
				\multirow{2}{*}{Dataset} & \multirow{2}{*}{SHOTS} & \multicolumn{3}{c}{ChatGPT-3.5} & \multicolumn{3}{c}{ERNIE-3.5} & \multicolumn{3}{c}{GPT-J} \\ \cline{3-11} 
				&  & F1 & P & R & F1 & P & R & F1 & P & R \\ \hline
				\multirow{4}{*}{LAP14} 
				& 1 & 49.74 & 47.40 & 52.31 & 25.88 & 25.35 & 26.43 & 46.07 & 47.28 & 44.92 \\
				& 5 & 52.09 & 48.26 & 56.56 & 23.91 & 22.22 & 25.87 & 54.28 & \textbf{54.58} & 53.97 \\
				& 10 & 52.39 & 48.65 & 56.75 & 23.62 & 23.06 & 24.21 & \underline{54.63} & 53.65 & 55.64 \\
				& 15 & 52.39 & 48.00 & \textbf{57.67} & 20.52 & 19.37 & 21.81 & \textbf{55.35} & \underline{53.85} & \underline{56.93} \\ \hline
				\multirow{4}{*}{RES14} 
				& 1 & 62.09 & 60.95 & 63.28 & 34.46 & 33.56 & 35.41 & 60.40 & 60.04 & 60.76 \\
				& 5 & 61.94 & 58.43 & 65.90 & 39.59 & 37.23 & 42.25 & \textbf{66.43} & \textbf{65.97} & 66.90 \\
				& 10 & 61.22 & 57.95 & 64.89 & 41.26 & 38.63 & 44.27 & 64.45 & 61.83 & \underline{67.30} \\
				& 15 & 62.26 & 58.61 & 66.40 & 37.13 & 35.58 & 38.83 & \underline{64.71} & \underline{61.97} & \textbf{67.71} \\ \hline
				\multirow{4}{*}{RES15} 
				& 1 & 52.63 & 55.84 & 49.77 & 28.44 & 25.75 & 31.75 & 48.79 & 44.62 & 53.82 \\
				& 5 & \underline{62.91} & \textbf{62.53} & 63.30 & 29.51 & 26.43 & 33.40 & 50.04 & 44.29 & 57.53 \\
				& 10 & \textbf{63.08} & \underline{60.07} & \underline{66.39} & 28.16 & 24.80 & 32.58 & 58.33 & 55.03 & 62.06 \\
				& 15 & 60.46 & 54.67 & \textbf{67.63} & 27.90 & 24.76 & 31.96 & 58.64 & 55.41 & 62.27 \\ \hline
				\multirow{4}{*}{RES16} 
				& 1 & 63.33 & \textbf{61.58} & 65.18 & 31.19 & 29.98 & 32.49 & 59.31 & 56.64 & 62.26 \\
				& 5 & 61.43 & 56.22 & 67.70 & 31.04 & 29.39 & 32.88 & 63.67 & 60.24 & 67.51 \\
				& 10 & \underline{63.99} & 58.10 & \underline{71.21} & 29.39 & 27.99 & 30.93 & 63.96 & \underline{60.77} & 67.51 \\
				& 15 & \textbf{65.44} & 59.71 & \textbf{72.37} & 29.96 & 28.40 & 31.71 & 63.16 & 59.18 & 67.70 \\ \hline
			\end{tabular}	
			}
		}
		\end{minipage}
	\end{table}
	
	\subsection{Compare with Baselines}
	To demonstrate the AiO model's advancement, we selected representative works from various sub-tasks as baseline models. These include models designed from both full-resource and low-resource perspectives. For fair comparison, we simulated a low-resource environment for full-resource models by using one-quarter of each dataset for training and testing on the same test set. For few-shot sentiment analysis models, we marked the backbone model and sample size. The selected baseline models are:
	\begin{enumerate}
		\item \textbf{ASGCN:} \citep{ALSC_3} A classic work designed for the ALSC task, which captures the internal dependencies of the text by introducing a GCN network.
		\item \textbf{DualGCN:} \citet{dualgcn} designed a dual graph convolution module to improve ALSC task performance and probabilistically refined traditional dependency parsing.
		\item \textbf{Span-ASTE:} \citep{span-aste} An early exploration of the ASTE task, which independently extracts entire entities using multi-word spans before pairing them.
		\item \textbf{T5-base:} \citet{NAPT} utilized a basic T5 module as the backbone model, providing baseline comparison results for various few-shot aspect-level sentiment analysis sub-tasks.
		\item \textbf{GTS-BERT:} \citet{gts} used BERT for feature extraction and enhances entity extraction robustness through a table labeling strategy.
		\item \textbf{GTS-BiLSTM:} \citet{gts} used BiLSTM for feature extraction and enhances entity extraction robustness through a table labeling strategy.
		\item \textbf{NAPT:} \citet{NAPT} enhanced the robustness of the T5 module in few-shot data through noisy ABSA pre-training and incorporates instruction tuning (IT) and multi-task learning (MTL) to improve model performance.
	\end{enumerate}
	
	\begin{table}[h]
	\centering
	\renewcommand{\arraystretch}{1}
	\setlength{\abovecaptionskip}{0pt}
	\setlength{\belowcaptionskip}{1pt}
	\begin{minipage}{\linewidth}
		\caption{Compare with Baselines on ALSC sub-task.}
		\label{baseline1}
		\resizebox{\columnwidth}{!}{
			\scalebox{1}{
			\begin{tabular}{@{}rccccc@{}}
				\toprule
				\multicolumn{2}{c}{Model}                      & LAP14 & RES14 & RES15 & RES16 \\ \midrule
				\multirow{3}{*}{ASGCN}                    & P  & 64.82 & 66.89 & 47.26 & 76.06 \\
				& R  & 65.73 & 67.05 & 51.35 & 60.30 \\
				& F1 & 64.14 & 66.44 & 49.04 & 63.90 \\ \midrule
				\multirow{3}{*}{DualGCN}                  & P  & 67.78 & 72.86 & 56.32 & 75.12 \\
				& R  & 63.44 & 64.96 & 60.17 & 67.58 \\
				& F1 & 63.93 & 67.31 & 58.18 & 71.15 \\ \midrule
				T5-base(5-shot)                           & F1 & -     & -     & 24.62 & 31.57 \\
				T5-base(10-shot)                          & F1 & -     & -     & 36.26 & 38.85 \\
				IT-T5(5-shot)                             & F1 & -     & -     & 24.75 & 36.26 \\
				IT-T5(10-shot)                            & F1 & -     & -     & 37.64 & 40.34 \\
				IT-MTL-NAPT-T5(5-shot)                    & F1 & -     & -     & 24.33 & 37.13 \\
				IT-MTL-NAPT-T5(10-shot)                   & F1 & -     & -     & 38.56 & 43.31 \\ \midrule
				\multirow{3}{*}{AiO-ChatGPT-3.5(5-shot)}  & P  & 68.43 & 76.39 & 71.69 & 75.78 \\
				& R  & 78.19 & 82.78 & 82.64 & 85.84 \\
				& F1 & 72.98 & 79.46 & 76.77 & 80.50 \\ \midrule
				\multirow{3}{*}{AiO-ChatGPT-3.5(10-shot)} & P  & 77.48 & 81.18 & 79.18 & 79.51 \\
				& R  & 82.51 & 85.97 & 85.42 & 86.73 \\
				& F1 & 79.92 & 83.51 & 82.18 & 82.96 \\ \bottomrule
			\end{tabular}%
			}
		}
	\end{minipage}
	\end{table}

	\begin{table}[h]
		\centering
		\renewcommand{\arraystretch}{1}
		\setlength{\abovecaptionskip}{0pt}
		\setlength{\belowcaptionskip}{1pt}
		\begin{minipage}{\linewidth}
			\caption{Compare with Baselines on AOPE sub-task.}
			\label{baseline2}
			\resizebox{\columnwidth}{!}{
				\scalebox{1}{
				\begin{tabular}{@{}rccccc@{}}
					\toprule
					\multicolumn{2}{c}{Model}                      & LAP14      & RES14      & RES15 & RES16      \\ \midrule
					\multirow{3}{*}{GTS-Bert}                 & P  & 61.79      & 73.18      & 65.14 & 67.06      \\
					& R  & 51.93      & 65.86      & 58.08 & 66.02      \\
					& F1 & 56.43      & 69.33      & 61.41 & 66.54      \\ \midrule
					\multirow{3}{*}{GTS-BiLSTM}               & P  & 64.31      & 70.83      & 70.23 & 71.16      \\
					& R  & 33.39      & 52.73      & 42.95 & 51.93      \\
					& F1 & 43.96      & 60.45      & 53.30 & 60.05      \\ \midrule
					T5-base(5-shot)                           & F1 & - & - & 39.44 & - \\
					T5-base(10-shot)                          & F1 & - & - & 49.28 & - \\
					IT-T5(5-shot)                             & F1 & - & - & 42.48 & - \\
					IT-T5(10-shot)                            & F1 & - & - & 51.03 & - \\
					IT-MTL-NAPT-T5(5-shot)                    & F1 & - & - & 39.59 & - \\
					IT-MTL-NAPT-T5(10-shot)                   & F1 & - & - & 50.34 & - \\ \midrule
					\multirow{3}{*}{AiO-ChatGPT-3.5(5-shot)}  & P  & 64.50      & 69.42      & 55.14 & 67.34      \\
					& R  & 60.44      & 70.12      & 66.39 & 71.40      \\
					& F1 & 62.41      & 69.77      & 60.24 & 69.31      \\ \midrule
					\multirow{3}{*}{AiO-ChatGPT-3.5(10-shot)} & P  & 64.47      & 70.72      & 55.10 & 67.52      \\
					& R  & 61.37      & 71.43      & 66.80 & 71.60      \\
					& F1 & 62.88      & 71.07      & 60.39 & 69.50      \\ \bottomrule
				\end{tabular}
				}
			}
		\end{minipage}
	\end{table}
	
	\begin{table}[h]
		\centering
		\renewcommand{\arraystretch}{1}
		\setlength{\abovecaptionskip}{0pt}
		\setlength{\belowcaptionskip}{1pt}
		\begin{minipage}{\linewidth}
			\caption{Compare with Baselines on ASTE sub-task.}
			\label{baseline3}
			\resizebox{\columnwidth}{!}{
				\scalebox{1}{
					\begin{tabular}{@{}rccccc@{}}
						\toprule
						\multicolumn{2}{c}{Model}                      & LAP14 & RES14      & RES15 & RES16 \\ \midrule
						\multirow{3}{*}{SPAN}                    & P  & 55.83 & 66.11      & 55.32 & 65.44 \\
						& R  & 43.00 & 58.67      & 52.33 & 64.21 \\
						& F1 & 48.58 & 62.17      & 53.78 & 64.82 \\ \midrule
						\multirow{3}{*}{GTS-Bert}                 & P  & 61.24 & 66.67      & 51.42 & 65.67 \\
						& R  & 43.49 & 61.01      & 44.38 & 59.46 \\
						& F1 & 50.86 & 63.71      & 47.64 & 62.41 \\ \midrule
						\multirow{3}{*}{GTS-BiLSTM}               & P  & 66.67 & 68.45      & 67.47 & 66.77 \\
						& R  & 27.89 & 45.15      & 39.88 & 42.66 \\
						& F1 & 39.33 & 54.41      & 50.13 & 52.06 \\ \midrule
						T5-base(5-shot)                           & F1 & 12.06 & - & 20.11 & 29.58 \\
						T5-base(10-shot)                         & F1 & 18.95 & - & 31.99 & 36.82 \\
						IT-T5(5-shot)                             & F1 & 28.14 & - & 24.44 & 34.10 \\
						IT-T5(10-shot)                            & F1 & 38.28 & - & 33.25 & 40.37 \\
						IT-MTL-NAPT-T5(5-shot)                    & F1 & 36.41 & - & 22.93 & 39.84 \\
						IT-MTL-NAPT-T5(10-shot)                   & F1 & 45.86 & - & 37.83 & 43.55 \\ \midrule
						\multirow{3}{*}{AiO-ChatGPT-3.5(5-shot)}  & P  & 54.58 & 65.97      & 44.29 & 60.24 \\
						& R  & 53.97 & 66.90      & 57.53 & 67.51 \\
						& F1 & 54.28 & 66.43      & 50.04 & 63.67 \\ \midrule
						\multirow{3}{*}{AiO-ChatGPT-3.5(10-shot)} & P  & 53.65 & 61.83      & 55.03 & 60.77 \\
						& R  & 55.64 & 67.30      & 62.06 & 67.51 \\
						& F1 & 54.63 & 64.45      & 58.33 & 63.96 \\ \bottomrule
					\end{tabular}
				}
			}
		\end{minipage}
	\end{table}
	The baseline model comparison results are shown in Table \ref{baseline1}, Table \ref{baseline2} and Table \ref{baseline3}. We used ChatGPT 3.5 with the proposed method for comparison. The results across various sub-tasks are highly consistent. Compared to traditional full-resource models, their performance drops sharply with limited training data, sometimes below models using LLMs as the backbone. For few-shot methods, the Heuristic-Enhanced Candidates Selection method we propose shows significant advantages over instruction tuning and multi-task learning methods.
	
	We analyzed the reasons for these results. Full-resource models perform poorly on few-shot tasks due to insufficient training data, preventing them from reaching their full potential. Traditional methods like instruction tuning depend on the model's generalization ability but fail to provide insightful understanding of downstream tasks. Our Heuristic-Enhanced Candidates Selection method improves LLMs' performance by generating rough answers first and then using LLMs' nuanced semantic understanding to refine them.
	
	\begin{figure*}[h]
		\centering
		\scalebox{.95}{
			\includegraphics{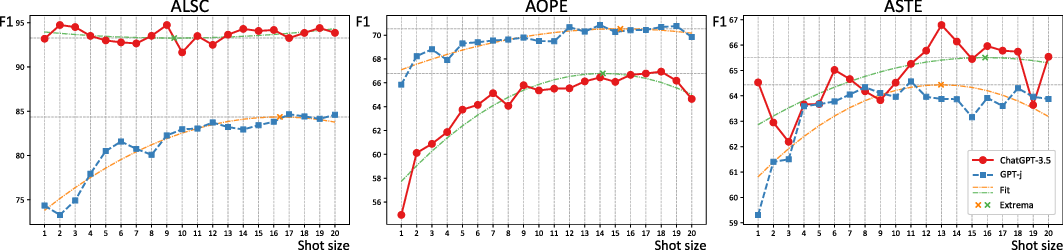}
		}
		\caption{Effect of shot size on results. Plotted as an example of ChatGPT-3.5 and GPT-J completing the ASTE task on top of RES16, with fitted curves set up for each fold and fitted extremes labeled.}
		\label{curve}
	\end{figure*}
						
	\section{Ablation Studies}
	\subsection{Impact of Shot Size}							
	The main experiment showed that the number of prompt samples affected the final results significantly. To investigate why, we selected ChatGPT-3.5 and GPT-J as test models and conducted an ablation experiment on shot size on the RES16 dataset. Specifically, we sequentially set the number of heuristic candidate samples (n-shots) from 1 to 20. A line chart was plotted using the sample quantity on the $x$ axis and the $F1$ score on the $y$ axis. We calculated a fitted curve for each line to depict the trend of result variations, marking the fitted extrema at corresponding positions. The results are depicted in Fig. \ref{curve}.

	By fitting the curve to the extreme of the curve, we found that ChatGPT-3.5 and GPT-J reached learning saturation from samples. ChatGPT-3.5 performance declined above a sample quantity of 15 for the ASTE task. Similarly, different tasks and models had distinct critical values, denoted by dashed lines. One cause for this could be that introducing heuristic samples to adjust the GPT model’s learning for downstream tasks guides the model to make better inferences with lighter prompt intensity, representing clearer semantic information. However, introducing too many prompt samples implies more complex semantics, which might lead to overfitting in local errors and cause a significant performance drop.
	
	\begin{table}[h]
		\centering
		\renewcommand{\arraystretch}{1}
		\setlength{\abovecaptionskip}{0pt}
		\setlength{\belowcaptionskip}{1pt}
		\begin{minipage}{\linewidth}
			\caption{Impact of the HCS strategy on results of the experiments on ChatGPT-3.5 and GPT-J for the LAP14 and RES16 datasets. The omission of HCS is denoted by "w/o HCS", and the decreasing values are calculated as the best results in the main experiments compared with the corresponding models and datasets.}
			\label{backbone}
			\resizebox{\columnwidth}{!}{
				\scalebox{1}{
				\begin{tabular}{@{}ccccc@{}}
				\toprule
				\multirow{2}{*}{Model} & \multirow{2}{*}{Dataset} & \multicolumn{3}{c}{Task} \\ \cmidrule(l){3-5} 
				&                          & ALSC   & AOPE   & ASTE   \\ \midrule
				\multirow{2}{*}{\begin{tabular}[c]{@{}c@{}}ChatGPT-3.5\\ w/o HCS\end{tabular}} 
				& LAP14                    & 84.44($\downarrow$1.31)   & 16.69($\downarrow$42.17)   & 25.64($\downarrow$26.75)   \\
				& RES16                    & 92.11($\downarrow$1.16)   & 30.03($\downarrow$36.21)   & 38.49($\downarrow$26.95)   \\ \midrule
				\multirow{2}{*}{\begin{tabular}[c]{@{}c@{}}GPT-j\\ w/o HCS\end{tabular}}            
				& LAP14                    & 50.57($\downarrow$29.60)   & 8.59($\downarrow$55.76)   & 7.15($\downarrow$48.2)   \\
				& RES16                    & 57.63($\downarrow$25.79)   & 18.57($\downarrow$51.70)   & 14.84($\downarrow$49.12)   \\ \bottomrule
				\end{tabular}
				}
			}
		\end{minipage}
	\end{table}
	
	\subsection{Impact of HCS strategy}
	Table \ref{backbone} demonstrates the role of the HCS strategy in the model. The results without the HCS strategy are denoted as "w/o HCS". We remove heuristic candidates from the prompt, retaining only the true results for each sample. For a quick analysis, we performed ablation experiments only on datasets LAP14 and RES16. It can be found that there is a significant impact on the overall performance without the HCS strategy, especially for GPT-J in the AOPE sub-task, where the maximum $F1$ score decrease was 55.76\%. It can be seen that the direct use of GPTs to process FSABSA performs poorly, while the introduction of the HCS strategy can boost the potential of GPTs for downstream sub-tasks.
	
	\begin{table}[h]
		\centering
		\renewcommand{\arraystretch}{1}
		\setlength{\abovecaptionskip}{0pt}
		\setlength{\belowcaptionskip}{1pt}
		\begin{minipage}{\columnwidth}
			\centering
			\caption{Comparison of the results of the dataset transfer experiments. Comparing the lift values for each dataset corresponding to the original candidate dataset, the rise results are highlighted in bold.}
			\label{transfer}
			\resizebox{0.8\columnwidth}{!}{
				\scalebox{1}{
				\begin{tabular}{@{}cccc@{}}
					\toprule
					Candidates              & Test & ChatGPT-3.5 & GPT-j \\ \midrule
					\multirow{3}{*}{\begin{tabular}[c]{@{}c@{}}RES16\\ 10-shots\end{tabular}}
					& LAP14    & 57.86(\textbf{$\uparrow$5.47})          & 56.59(\textbf{$\uparrow$1.96})   \\
					& RES14    & 67.96(\textbf{$\uparrow$6.74})          & 67.12(\textbf{$\uparrow$2.67})  \\
					& RES15    & 63.33(\textbf{$\uparrow$0.25})          & 60.40(\textbf{$\uparrow$2.07})  \\ \midrule
					\multirow{3}{*}{\begin{tabular}[c]{@{}c@{}}LAP14\\ 10-shots\end{tabular}} 	
					& RES14    & 65.85(\textbf{$\uparrow$4.63})          & 63.44($\downarrow$1.01)  \\
					& RES15    & 53.84($\downarrow$9.24)          & 52.21($\downarrow$6.12)  \\
					& RES16    & 64.82($\downarrow$0.83)          & 59.89($\downarrow$4.07)  \\ \bottomrule
				\end{tabular}
				}
			}
		\end{minipage}
	\end{table}
	
	\subsection{Transfer experiment}
	As is well known, FSABSA is constrained by insufficient datasets. To evaluate the AiO model's transferability across datasets, we conducted transfer tests. In the first stage, we trained the backbone model using one dataset. In the second stage, we used this trained backbone to assist in reasoning on the test dataset. We chose the challenging ASTE task for comparison. The experimental results are detailed in Table \ref{transfer}, with the training dataset in the “Candidates” column and the test dataset in the “Test” column. This experiment design aims to train prompt candidates using only one dataset. Within the same domain (restaurant or laptop), different datasets have the same evaluation subject but differ in expression. Across domains, datasets have different evaluation subjects but share common evaluative statements. This design helps reflect the AiO model's performance differences in sentiment expression semantics and syntax.
	
	\section{Conclusion}
	This paper proposed addressing the multifaceted challenges of FSABSA by introducing the All in One (AiO) model. We proposed a two-stage approach leveraging a specific backbone network and GPTs. Our work marks the first endeavor to employ GPTs for FSABSA, while defining a set of generalized learning paradigms that can be repeated. Extensive experiments on diverse datasets demonstrated AiO’s robust generalization and notable contributions. It will provide valuable insights that will guide future development in the intersection of GPT models and sentiment analysis, particularly in scenarios with limited data availability.	This work has several limitations. On the one hand, the HCS strategy requires selecting a suitable pre-trained model for generating candidates, which potentially increases the risk of computing power spend. On the other hand, although we conducted a large number of experiments to validate the effect, due to the limitation of space, our experiments can only select some models for comparison and it is difficult to cover all GPTs models.  

\section*{Acknowledgements}
This work was supported by the key project of the National Natural Science Foundation of China under Grant 62137001.

\section*{Declaration of competing interest}
The authors declare that they have no known competing financial interests or personal relationships that could have appeared to influence the work reported in this paper.

\section*{CRediT authorship contribution statement}
\textbf{Baoxing Jiang:} Conceptualization, Methodology, Software, Validation, Writing - Original Draft. \textbf{Yujie Wan:} Writing - Review \& Editing, Formal analysis. \textbf{Shenggen Ju:} Funding acquisition, Project administration.

\section*{Data availability}
The link to all data and code are available in the manuscript.






\begin{thebibliography}{46}
	\expandafter\ifx\csname natexlab\endcsname\relax\def\natexlab#1{#1}\fi
	\providecommand{\url}[1]{\texttt{#1}}
	\providecommand{\href}[2]{#2}
	\providecommand{\path}[1]{#1}
	\providecommand{\DOIprefix}{ }
	\providecommand{\ArXivprefix}{arXiv:}
	\providecommand{\URLprefix}{URL: }
	\providecommand{\Pubmedprefix}{pmid:}
	\providecommand{\doi}[1]{\href{http://dx.doi.org/#1}{\path{#1}}}
	\providecommand{\Pubmed}[1]{\href{pmid:#1}{\path{#1}}}
	\providecommand{\bibinfo}[2]{#2}
	\ifx\xfnm\relax \def\xfnm[#1]{\unskip,\space#1}\fi
	\bibitem[{Brown et~al.(2020)Brown, Mann, Ryder, Subbiah, Kaplan, Dhariwal,
		Neelakantan, Shyam, Sastry, Askell, Agarwal, Herbert-Voss, Krueger, Henighan,
		Child, Ramesh, Ziegler, Wu, Winter, Hesse, Chen, Sigler, Litwin, Gray, Chess,
		Clark, Berner, McCandlish, Radford, Sutskever \& Amodei}]{GPT3}
	\bibinfo{author}{Brown, T.}, \bibinfo{author}{Mann, B.},
	\bibinfo{author}{Ryder, N.}, \bibinfo{author}{Subbiah, M.},
	\bibinfo{author}{Kaplan, J.~D.}, \bibinfo{author}{Dhariwal, P.},
	\bibinfo{author}{Neelakantan, A.}, \bibinfo{author}{Shyam, P.},
	\bibinfo{author}{Sastry, G.}, \bibinfo{author}{Askell, A.},
	\bibinfo{author}{Agarwal, S.}, \bibinfo{author}{Herbert-Voss, A.},
	\bibinfo{author}{Krueger, G.}, \bibinfo{author}{Henighan, T.},
	\bibinfo{author}{Child, R.}, \bibinfo{author}{Ramesh, A.},
	\bibinfo{author}{Ziegler, D.}, \bibinfo{author}{Wu, J.},
	\bibinfo{author}{Winter, C.}, \bibinfo{author}{Hesse, C.},
	\bibinfo{author}{Chen, M.}, \bibinfo{author}{Sigler, E.},
	\bibinfo{author}{Litwin, M.}, \bibinfo{author}{Gray, S.},
	\bibinfo{author}{Chess, B.}, \bibinfo{author}{Clark, J.},
	\bibinfo{author}{Berner, C.}, \bibinfo{author}{McCandlish, S.},
	\bibinfo{author}{Radford, A.}, \bibinfo{author}{Sutskever, I.}, \&
	\bibinfo{author}{Amodei, D.}(\textcharter{\bibinfo{year}{2020}}).
	\newblock \bibinfo{title}{Language models are few-shot learners}.
	\newblock In {\it \bibinfo{booktitle}{Advances in Neural Information Processing
			Systems 33, NeurIPS, December 6-12, 2020}\/} (pp.
	\bibinfo{pages}{1877--1901}).
	\newblock \bibinfo{address}{Vancouver, BC, Canada}: \bibinfo{publisher}{Curran
		Associates Inc.} volume~\bibinfo{volume}{33}.
	\newblock \URLprefix
	\url{https://proceedings.neurips.cc/paper_files/paper/2020/file/1457c0d6bfcb4967418bfb8ac142f64a-Paper.pdf}.
	\bibitem[{Das \& Chen(2001)}]{cucaoabsa1}
	\bibinfo{author}{Das, S.~R.}, \& \bibinfo{author}{Chen,
		M.~Y.}(\textcharter{\bibinfo{year}{2001}}).
	\newblock \bibinfo{title}{Yahoo! for amazon: Sentiment parsing from small talk
		on the web}.
	\newblock In {\it \bibinfo{booktitle}{EFA 2001 Barcelona Meetings, Available at
			SSRN, {EFA}, August 5, 2001}\/} (p.~\bibinfo{pages}{45}).
	\newblock \bibinfo{address}{Bangkok, Thailand}: \bibinfo{publisher}{Elsevier}.
	\newblock \DOIprefix\doi{10.2139/ssrn.276189}.
	\bibitem[{Devlin et~al.(2019)Devlin, Chang, Lee \& Toutanova}]{bert}
	\bibinfo{author}{Devlin, J.}, \bibinfo{author}{Chang, M.},
	\bibinfo{author}{Lee, K.}, \& \bibinfo{author}{Toutanova,
		K.}(\textcharter{\bibinfo{year}{2019}}).
	\newblock \bibinfo{title}{{BERT:} pre-training of deep bidirectional
		transformers for language understanding}.
	\newblock In {\it \bibinfo{booktitle}{Proceedings of the 2019 Conference of the
			North American Chapter of the Association for Computational Linguistics:
			Human Language Technologies, {NAACL-HLT}}\/} (pp.
	\bibinfo{pages}{4171--4186}).
	\newblock \bibinfo{address}{Minneapolis, MN, USA}: \bibinfo{publisher}{ACL}.
	\newblock \DOIprefix\doi{10.18653/V1/N19-1423}.
	\bibitem[{Dong et~al.(2023)Dong, Xu, Jiang, Li \& Liu}]{jbxfewshot}
	\bibinfo{author}{Dong, K.}, \bibinfo{author}{Xu, F.}, \bibinfo{author}{Jiang,
		B.}, \bibinfo{author}{Li, H.}, \& \bibinfo{author}{Liu,
		P.}(\textcharter{\bibinfo{year}{2023}}).
	\newblock \bibinfo{title}{Multitask-based cluster transmission for few-shot
		text classification}.
	\newblock In {\it \bibinfo{booktitle}{Knowledge Science, Engineering and
			Management - 16th International Conference, {KSEM}}\/} (pp.
	\bibinfo{pages}{66--77}).
	\newblock \bibinfo{address}{Guangzhou, China}: \bibinfo{publisher}{Springer}
	volume \bibinfo{volume}{14117}.
	\newblock \DOIprefix\doi{10.1007/978-3-031-40283-8\_7}.
	\bibitem[{Fan et~al.(2019{\natexlab{a}})Fan, Wu, Dai, Huang \& Chen}]{AOPE_1}
	\bibinfo{author}{Fan, Z.}, \bibinfo{author}{Wu, Z.}, \bibinfo{author}{Dai, X.},
	\bibinfo{author}{Huang, S.}, \& \bibinfo{author}{Chen,
		J.}(\textcharter{\bibinfo{year}{2019}{\natexlab{a}}}).
	\newblock \bibinfo{title}{Target-oriented opinion words extraction with
		target-fused neural sequence labeling}.
	\newblock In {\it \bibinfo{booktitle}{Proceedings of the 2019 Conference of the
			North American Chapter of the Association for Computational {NAACL-HLT}}\/}
	(pp. \bibinfo{pages}{2509--2518}).
	\newblock \bibinfo{address}{Minneapolis, MN, USA}: \bibinfo{publisher}{ACL}.
	\newblock \DOIprefix\doi{10.18653/V1/N19-1259}.
	\bibitem[{Fan et~al.(2019{\natexlab{b}})Fan, Wu, Dai, Huang \&
		Chen}]{fan-first-label-dataset}
	\bibinfo{author}{Fan, Z.}, \bibinfo{author}{Wu, Z.}, \bibinfo{author}{Dai, X.},
	\bibinfo{author}{Huang, S.}, \& \bibinfo{author}{Chen,
		J.}(\textcharter{\bibinfo{year}{2019}{\natexlab{b}}}).
	\newblock \bibinfo{title}{Target-oriented opinion words extraction with
		target-fused neural sequence labeling}.
	\newblock In {\it \bibinfo{booktitle}{Proceedings of the 2019 Conference of the
			North American Chapter of the Association for Computational Linguistics
			(NAACL)}\/} (pp. \bibinfo{pages}{2509--2518}).
	\newblock \bibinfo{address}{Minneapolis, MN, USA}: \bibinfo{publisher}{ACL}.
	\newblock \DOIprefix\doi{10.18653/v1/n19-1259}.
	\bibitem[{Fei et~al.(2022)Fei, Zeng, Zhao, Li \& Xiao}]{fewshotTE}
	\bibinfo{author}{Fei, J.}, \bibinfo{author}{Zeng, W.}, \bibinfo{author}{Zhao,
		X.}, \bibinfo{author}{Li, X.}, \& \bibinfo{author}{Xiao,
		W.}(\textcharter{\bibinfo{year}{2022}}).
	\newblock \bibinfo{title}{Few-shot relational triple extraction with
		perspective transfer network}.
	\newblock In {\it \bibinfo{booktitle}{Proceedings of the 31st {ACM}
			International Conference on Information {\&} Knowledge Management, {CIKM}}\/}
	(pp. \bibinfo{pages}{488--498}).
	\newblock \bibinfo{address}{Atlanta, GA, USA}: \bibinfo{publisher}{ACM}.
	\newblock \DOIprefix\doi{10.1145/3511808.3557323}.
	\bibitem[{Ganu et~al.(2009)Ganu, Elhadad \& Marian}]{MAMS_cite}
	\bibinfo{author}{Ganu, G.}, \bibinfo{author}{Elhadad, N.}, \&
	\bibinfo{author}{Marian, A.}(\textcharter{\bibinfo{year}{2009}}).
	\newblock \bibinfo{title}{Beyond the stars: Improving rating predictions using
		review text content}.
	\newblock In {\it \bibinfo{booktitle}{12th International Workshop on the Web
			and Databases, {WebDB}, June 28}\/} (pp. \bibinfo{pages}{1--6}).
	\newblock \bibinfo{address}{Providence, Rhode Island, USA}:
	\bibinfo{publisher}{ACM}.
	\newblock \URLprefix
	\url{http://webdb09.cse.buffalo.edu/papers/Paper9/WebDB.pdf}.
	\bibitem[{He et~al.(2018)He, Lee, Ng \& Dahlmeier}]{ALSC_2}
	\bibinfo{author}{He, R.}, \bibinfo{author}{Lee, W.~S.}, \bibinfo{author}{Ng,
		H.~T.}, \& \bibinfo{author}{Dahlmeier,
		D.}(\textcharter{\bibinfo{year}{2018}}).
	\newblock \bibinfo{title}{Effective attention modeling for aspect-level
		sentiment classification}.
	\newblock In {\it \bibinfo{booktitle}{Proceedings of the 27th International
			Conference on Computational Linguistics, {COLING}, August 20-26, 2018}\/}
	(pp. \bibinfo{pages}{1121--1131}).
	\newblock \bibinfo{address}{Santa Fe, New Mexico, USA}:
	\bibinfo{publisher}{ACL}.
	\bibitem[{Huang et~al.(2023)Huang, Peng, Liu, Yang, Wang,
		Orellana{-}Mart{\'{\i}}n \& P{\'{e}}rez{-}Jim{\'{e}}nez}]{alsc2}
	\bibinfo{author}{Huang, Y.}, \bibinfo{author}{Peng, H.}, \bibinfo{author}{Liu,
		Q.}, \bibinfo{author}{Yang, Q.}, \bibinfo{author}{Wang, J.},
	\bibinfo{author}{Orellana{-}Mart{\'{\i}}n, D.}, \&
	\bibinfo{author}{P{\'{e}}rez{-}Jim{\'{e}}nez,
		M.~J.}(\textcharter{\bibinfo{year}{2023}}).
	\newblock \bibinfo{title}{Attention-enabled gated spiking neural p model for
		aspect-level sentiment classification}.
	\newblock {\it \bibinfo{journal}{Neural Networks}\/},  {\it
		\bibinfo{volume}{157}\/}, \bibinfo{pages}{437--443}.
	\DOIprefix\doi{10.1016/j.neunet.2022.11.006}.
	\bibitem[{Jiang et~al.(2023{\natexlab{a}})Jiang, Liang, Liu, Dong \&
		Li}]{jbxaste}
	\bibinfo{author}{Jiang, B.}, \bibinfo{author}{Liang, S.}, \bibinfo{author}{Liu,
		P.}, \bibinfo{author}{Dong, K.}, \& \bibinfo{author}{Li,
		H.}(\textcharter{\bibinfo{year}{2023}{\natexlab{a}}}).
	\newblock \bibinfo{title}{A semantically enhanced dual encoder for aspect
		sentiment triplet extraction}.
	\newblock {\it \bibinfo{journal}{Neurocomputing}\/},  {\it
		\bibinfo{volume}{562}\/}, \bibinfo{pages}{126917}.
	\DOIprefix\doi{10.1016/J.NEUCOM.2023.126917}.
	\bibitem[{Jiang et~al.(2023{\natexlab{b}})Jiang, Xu \& Liu}]{jbxalsc}
	\bibinfo{author}{Jiang, B.}, \bibinfo{author}{Xu, G.}, \& \bibinfo{author}{Liu,
		P.}(\textcharter{\bibinfo{year}{2023}{\natexlab{b}}}).
	\newblock \bibinfo{title}{Aspect-level sentiment classification via location
		enhanced aspect-merged graph convolutional networks}.
	\newblock {\it \bibinfo{journal}{The Journal of Supercomputing}\/},  {\it
		\bibinfo{volume}{79}\/}, \bibinfo{pages}{9666--9691}.
	\DOIprefix\doi{10.1007/S11227-022-05002-4}.
	\bibitem[{Jiang et~al.(2019)Jiang, Chen, Xu, Ao \& Yang}]{MAMS}
	\bibinfo{author}{Jiang, Q.}, \bibinfo{author}{Chen, L.}, \bibinfo{author}{Xu,
		R.}, \bibinfo{author}{Ao, X.}, \& \bibinfo{author}{Yang,
		M.}(\textcharter{\bibinfo{year}{2019}}).
	\newblock \bibinfo{title}{A challenge dataset and effective models for
		aspect-based sentiment analysis}.
	\newblock In {\it \bibinfo{booktitle}{Proceedings of the 2019 Conference on
			Empirical Methods in Natural Language Processing and the 9th International
			Joint Conference on Natural Language Processing, {EMNLP-IJCNLP}, November
			3-7, 2019}\/} (pp. \bibinfo{pages}{6279--6284}).
	\newblock \bibinfo{address}{Hong Kong, China}: \bibinfo{publisher}{ACL}.
	\newblock \DOIprefix\doi{10.18653/V1/D19-1654}.
	\bibitem[{Kang et~al.(2022)Kang, Kim, Yun, Lee \& Jung}]{aoe1}
	\bibinfo{author}{Kang, T.}, \bibinfo{author}{Kim, S.}, \bibinfo{author}{Yun,
		H.}, \bibinfo{author}{Lee, H.}, \& \bibinfo{author}{Jung,
		K.}(\textcharter{\bibinfo{year}{2022}}).
	\newblock \bibinfo{title}{Gated relational encoder-decoder model for
		target-oriented opinion word extraction}.
	\newblock {\it \bibinfo{journal}{{IEEE} Access}\/},  {\it
		\bibinfo{volume}{10}\/}, \bibinfo{pages}{130507--130517}.
	\DOIprefix\doi{10.1109/ACCESS.2022.3228835}.
	\bibitem[{Li et~al.(2021)Li, Chen, Feng, Ma, Wang \& Hovy}]{dualgcn}
	\bibinfo{author}{Li, R.}, \bibinfo{author}{Chen, H.}, \bibinfo{author}{Feng,
		F.}, \bibinfo{author}{Ma, Z.}, \bibinfo{author}{Wang, X.}, \&
	\bibinfo{author}{Hovy, E.~H.}(\textcharter{\bibinfo{year}{2021}}).
	\newblock \bibinfo{title}{Dual graph convolutional networks for aspect-based
		sentiment analysis}.
	\newblock In {\it \bibinfo{booktitle}{Proceedings of the 59th Annual Meeting of
			the Association for Computational Linguistics and the 11th International
			Joint Conference on Natural Language Processing, {ACL/IJCNLP}}\/} (pp.
	\bibinfo{pages}{6319--6329}).
	\newblock \bibinfo{address}{Virtual Event}: \bibinfo{publisher}{Association for
		Computational Linguistics}.
	\bibitem[{Liang et~al.(2023)Liang, Wei, Mao, Fu, Fang \& Chen}]{ASTE_1}
	\bibinfo{author}{Liang, S.}, \bibinfo{author}{Wei, W.}, \bibinfo{author}{Mao,
		X.}, \bibinfo{author}{Fu, Y.}, \bibinfo{author}{Fang, R.}, \&
	\bibinfo{author}{Chen, D.}(\textcharter{\bibinfo{year}{2023}}).
	\newblock \bibinfo{title}{Stage: Span tagging and greedy inference scheme for
		aspect sentiment triplet extraction}.
	\newblock In {\it \bibinfo{booktitle}{Thirty-Seventh AAAI Conference on
			Artificial Intelligence, {AAAI}, Thirty-Fifth Conference on Innovative
			Applications of Artificial Intelligence, {IAAI}, Thirteenth Symposium on
			Educational Advances in Artificial Intelligence, {EAAI}, February 7-14,
			2023}\/} (pp. \bibinfo{pages}{13174--13182}).
	\newblock \bibinfo{address}{Washington, DC, USA}: \bibinfo{publisher}{AAAI
		Press}.
	\newblock \DOIprefix\doi{10.1609/aaai.v37i11.26547}.
	\bibitem[{Liu et~al.(2015)Liu, Joty \& Meng}]{ae1}
	\bibinfo{author}{Liu, P.}, \bibinfo{author}{Joty, S.~R.}, \&
	\bibinfo{author}{Meng, H.~M.}(\textcharter{\bibinfo{year}{2015}}).
	\newblock \bibinfo{title}{Fine-grained opinion mining with recurrent neural
		networks and word embeddings}.
	\newblock In {\it \bibinfo{booktitle}{Proceedings of the 2015 Conference on
			Empirical Methods in Natural Language Processing, {EMNLP}, September 17-21,
			2015}\/} (pp. \bibinfo{pages}{1433--1443}).
	\newblock \bibinfo{address}{Lisbon, Portugal}: \bibinfo{publisher}{ACL}.
	\newblock \DOIprefix\doi{10.18653/V1/D15-1168}.
	\bibitem[{Liu et~al.(2023)Liu, Yuan, Fu, Jiang, Hayashi \& Neubig}]{now1}
	\bibinfo{author}{Liu, P.}, \bibinfo{author}{Yuan, W.}, \bibinfo{author}{Fu,
		J.}, \bibinfo{author}{Jiang, Z.}, \bibinfo{author}{Hayashi, H.}, \&
	\bibinfo{author}{Neubig, G.}(\textcharter{\bibinfo{year}{2023}}).
	\newblock \bibinfo{title}{Pre-train, prompt, and predict: {A} systematic survey
		of prompting methods in natural language processing}.
	\newblock {\it \bibinfo{journal}{{ACM} Comput. Surv.}\/},  {\it
		\bibinfo{volume}{55}\/}, \bibinfo{pages}{195:1--195:35}.
	\DOIprefix\doi{10.1145/3560815}.
	\bibitem[{Liu et~al.(2022)Liu, Li, Fei \& Ji}]{aoe2}
	\bibinfo{author}{Liu, Y.}, \bibinfo{author}{Li, F.}, \bibinfo{author}{Fei, H.},
	\& \bibinfo{author}{Ji, D.}(\textcharter{\bibinfo{year}{2022}}).
	\newblock \bibinfo{title}{Pair-wise aspect and opinion terms extraction as
		graph parsing via a novel mutually-aware interaction mechanism}.
	\newblock {\it \bibinfo{journal}{Neurocomputing}\/},  {\it
		\bibinfo{volume}{493}\/}, \bibinfo{pages}{268--280}.
	\DOIprefix\doi{10.1016/j.neucom.2022.04.064}.
	\bibitem[{Morinaga et~al.(2002)Morinaga, Yamanishi, Tateishi \&
		Fukushima}]{cucaoabsa2}
	\bibinfo{author}{Morinaga, S.}, \bibinfo{author}{Yamanishi, K.},
	\bibinfo{author}{Tateishi, K.}, \& \bibinfo{author}{Fukushima,
		T.}(\textcharter{\bibinfo{year}{2002}}).
	\newblock \bibinfo{title}{Mining product reputations on the web}.
	\newblock In {\it \bibinfo{booktitle}{Proceedings of the Eighth {ACM} {SIGKDD}
			International Conference on Knowledge Discovery and Data Mining, {KDD}, July
			23-26, 2002}\/} (pp. \bibinfo{pages}{341--349}).
	\newblock \bibinfo{address}{Edmonton, Alberta, Canada}:
	\bibinfo{publisher}{ACM}.
	\newblock \DOIprefix\doi{10.1145/775047.775098}.
	\bibitem[{OpenAI(2023)}]{GPT4}
	\bibinfo{author}{OpenAI}(\textcharter{\bibinfo{year}{2023}}).
	\newblock \bibinfo{title}{Gpt-4 technical report}.
	\newblock \href{http://arxiv.org/abs/2303.08774}{\tt arXiv:2303.08774}.
	\bibitem[{Pang \& Lee(2008)}]{absa_beign}
	\bibinfo{author}{Pang, B.}, \& \bibinfo{author}{Lee,
		L.}(\textcharter{\bibinfo{year}{2008}}).
	\newblock \bibinfo{title}{Opinion mining and sentiment analysis}.
	\newblock {\it \bibinfo{journal}{Foundations and Trends in Information
			Retrieval}\/},  {\it \bibinfo{volume}{2}\/}, \bibinfo{pages}{1--135}.
	\DOIprefix\doi{10.1561/1500000011}.
	\bibitem[{Peng et~al.(2020)Peng, Xu, Bing, Huang, Lu \& Si}]{peng-two-stage}
	\bibinfo{author}{Peng, H.}, \bibinfo{author}{Xu, L.}, \bibinfo{author}{Bing,
		L.}, \bibinfo{author}{Huang, F.}, \bibinfo{author}{Lu, W.}, \&
	\bibinfo{author}{Si, L.}(\textcharter{\bibinfo{year}{2020}}).
	\newblock \bibinfo{title}{Knowing what, how and why: A near complete solution
		for aspect-based sentiment analysis}.
	\newblock In {\it \bibinfo{booktitle}{The Thirty-Fourth {AAAI} Conference on
			Artificial Intelligence (AAAI)}\/} (pp. \bibinfo{pages}{8600--8607}).
	\newblock \bibinfo{address}{New York, NY, USA}: \bibinfo{publisher}{{AAAI}
		Press}.
	\newblock \DOIprefix\doi{10.1609/aaai.v34i05.6383}.
	\bibitem[{Pontiki et~al.(2016)Pontiki, Galanis, Papageorgiou, Androutsopoulos,
		Manandhar, Al{-}Smadi, Al{-}Ayyoub, Zhao, Qin, Clercq, Hoste, Apidianaki,
		Tannier, Loukachevitch, Kotelnikov, Bel, Zafra \&
		Eryigit}]{semeval-2016-task5}
	\bibinfo{author}{Pontiki, M.}, \bibinfo{author}{Galanis, D.},
	\bibinfo{author}{Papageorgiou, H.}, \bibinfo{author}{Androutsopoulos, I.},
	\bibinfo{author}{Manandhar, S.}, \bibinfo{author}{Al{-}Smadi, M.},
	\bibinfo{author}{Al{-}Ayyoub, M.}, \bibinfo{author}{Zhao, Y.},
	\bibinfo{author}{Qin, B.}, \bibinfo{author}{Clercq, O.~D.},
	\bibinfo{author}{Hoste, V.}, \bibinfo{author}{Apidianaki, M.},
	\bibinfo{author}{Tannier, X.}, \bibinfo{author}{Loukachevitch, N.~V.},
	\bibinfo{author}{Kotelnikov, E.~V.}, \bibinfo{author}{Bel, N.},
	\bibinfo{author}{Zafra, S. M.~J.}, \& \bibinfo{author}{Eryigit,
		G.}(\textcharter{\bibinfo{year}{2016}}).
	\newblock \bibinfo{title}{Semeval-2016 task 5: Aspect based sentiment
		analysis}.
	\newblock In {\it \bibinfo{booktitle}{Proceedings of the 10th International
			Workshop on Semantic Evaluation (SemEval{@}NAACL-HLT)}\/} (pp.
	\bibinfo{pages}{19--30}).
	\newblock \bibinfo{address}{San Diego, CA, USA}: \bibinfo{publisher}{ACL}.
	\newblock \DOIprefix\doi{10.18653/v1/s16-1002}.
	\bibitem[{Pontiki et~al.(2015)Pontiki, Galanis, Papageorgiou, Manandhar \&
		Androutsopoulos}]{semeval-2015-task12}
	\bibinfo{author}{Pontiki, M.}, \bibinfo{author}{Galanis, D.},
	\bibinfo{author}{Papageorgiou, H.}, \bibinfo{author}{Manandhar, S.}, \&
	\bibinfo{author}{Androutsopoulos, I.}(\textcharter{\bibinfo{year}{2015}}).
	\newblock \bibinfo{title}{Semeval-2015 task 12: Aspect based sentiment
		analysis}.
	\newblock In {\it \bibinfo{booktitle}{Proceedings of the 9th International
			Workshop on Semantic Evaluation (SemEval{@}NAACL-HLT)}\/} (pp.
	\bibinfo{pages}{486--495}).
	\newblock \bibinfo{address}{Denver, Colorado, USA}: \bibinfo{publisher}{ACL}.
	\newblock \DOIprefix\doi{10.18653/v1/s15-2082}.
	\bibitem[{Pontiki et~al.(2014)Pontiki, Galanis, Pavlopoulos, Papageorgiou,
		Androutsopoulos \& Manandhar}]{semeval-2014-task4}
	\bibinfo{author}{Pontiki, M.}, \bibinfo{author}{Galanis, D.},
	\bibinfo{author}{Pavlopoulos, J.}, \bibinfo{author}{Papageorgiou, H.},
	\bibinfo{author}{Androutsopoulos, I.}, \& \bibinfo{author}{Manandhar,
		S.}(\textcharter{\bibinfo{year}{2014}}).
	\newblock \bibinfo{title}{Semeval-2014 task 4: Aspect based sentiment
		analysis}.
	\newblock In {\it \bibinfo{booktitle}{Proceedings of the 8th International
			Workshop on Semantic Evaluation (SemEval{@}COLING)}\/} (pp.
	\bibinfo{pages}{27--35}).
	\newblock \bibinfo{address}{Dublin, Ireland}: \bibinfo{publisher}{ACL}.
	\newblock \DOIprefix\doi{10.3115/v1/s14-2004}.
	\bibitem[{Shao et~al.(2023)Shao, Yu, Wang \& Yu}]{kbvqa2}
	\bibinfo{author}{Shao, Z.}, \bibinfo{author}{Yu, Z.}, \bibinfo{author}{Wang,
		M.}, \& \bibinfo{author}{Yu, J.}(\textcharter{\bibinfo{year}{2023}}).
	\newblock \bibinfo{title}{Prompting large language models with answer
		heuristics for knowledge-based visual question answering}.
	\newblock In {\it \bibinfo{booktitle}{{IEEE/CVF} Conference on Computer Vision
			and Pattern Recognition, {CVPR}, June 17-24, 2023}\/} (pp.
	\bibinfo{pages}{14974--14983}).
	\newblock \bibinfo{address}{Vancouver, BC, Canada}: \bibinfo{publisher}{IEEE}.
	\newblock \DOIprefix\doi{10.1109/CVPR52729.2023.01438}.
	\bibitem[{Sun et~al.(2024)Sun, Zhang, Liu, Bao \& Chen}]{harnessing}
	\bibinfo{author}{Sun, X.}, \bibinfo{author}{Zhang, K.}, \bibinfo{author}{Liu,
		Q.}, \bibinfo{author}{Bao, M.}, \& \bibinfo{author}{Chen,
		Y.}(\textcharter{\bibinfo{year}{2024}}).
	\newblock \bibinfo{title}{Harnessing domain insights: A prompt knowledge tuning
		method for aspect-based sentiment analysis}.
	\newblock {\it \bibinfo{journal}{Knowledge-Based Systems}\/},  (p.
	\bibinfo{pages}{111975}).
	\bibitem[{Sun et~al.(2021)Sun, Wang, Feng, Ding, Pang, Shang, Liu, Chen, Zhao,
		Lu, Liu, Wu, Gong, Liang, Shang, Sun, Liu, Ouyang, Yu, Tian, Wu \&
		Wang}]{ERNIE}
	\bibinfo{author}{Sun, Y.}, \bibinfo{author}{Wang, S.}, \bibinfo{author}{Feng,
		S.}, \bibinfo{author}{Ding, S.}, \bibinfo{author}{Pang, C.},
	\bibinfo{author}{Shang, J.}, \bibinfo{author}{Liu, J.},
	\bibinfo{author}{Chen, X.}, \bibinfo{author}{Zhao, Y.}, \bibinfo{author}{Lu,
		Y.}, \bibinfo{author}{Liu, W.}, \bibinfo{author}{Wu, Z.},
	\bibinfo{author}{Gong, W.}, \bibinfo{author}{Liang, J.},
	\bibinfo{author}{Shang, Z.}, \bibinfo{author}{Sun, P.}, \bibinfo{author}{Liu,
		W.}, \bibinfo{author}{Ouyang, X.}, \bibinfo{author}{Yu, D.},
	\bibinfo{author}{Tian, H.}, \bibinfo{author}{Wu, H.}, \&
	\bibinfo{author}{Wang, H.}(\textcharter{\bibinfo{year}{2021}}).
	\newblock \bibinfo{title}{Ernie 3.0: Large-scale knowledge enhanced
		pre-training for language understanding and generation}.
	\newblock \href{http://arxiv.org/abs/2107.02137}{\tt arXiv:2107.02137}.
	\bibitem[{Tang et~al.(2016)Tang, Qin \& Liu}]{ALSC_1}
	\bibinfo{author}{Tang, D.}, \bibinfo{author}{Qin, B.}, \& \bibinfo{author}{Liu,
		T.}(\textcharter{\bibinfo{year}{2016}}).
	\newblock \bibinfo{title}{Aspect level sentiment classification with deep
		memory network}.
	\newblock In {\it \bibinfo{booktitle}{Proceedings of the 2016 Conference on
			Empirical Methods in Natural Language Processing, {EMNLP}, November 1-4,
			2016}\/} (pp. \bibinfo{pages}{214--224}).
	\newblock \bibinfo{address}{Austin, Texas, USA}: \bibinfo{publisher}{ACL}.
	\newblock \DOIprefix\doi{10.18653/V1/D16-1021}.
	\bibitem[{Vacareanu et~al.(2024)Vacareanu, Varia, Halder, Wang, Paolini, John,
		Ballesteros \& Muresan}]{NAPT}
	\bibinfo{author}{Vacareanu, R.}, \bibinfo{author}{Varia, S.},
	\bibinfo{author}{Halder, K.}, \bibinfo{author}{Wang, S.},
	\bibinfo{author}{Paolini, G.}, \bibinfo{author}{John, N.~A.},
	\bibinfo{author}{Ballesteros, M.}, \& \bibinfo{author}{Muresan,
		S.}(\textcharter{\bibinfo{year}{2024}}).
	\newblock \bibinfo{title}{A weak supervision approach for few-shot aspect based
		sentiment analysis}.
	\newblock In {\it \bibinfo{booktitle}{Proceedings of the 18th Conference of the
			European Chapter of the Association for Computational Linguistics, {EACL}}\/}
	(pp. \bibinfo{pages}{2734--2752}).
	\newblock \bibinfo{address}{St. Julian's, Malta}: \bibinfo{publisher}{ACL}.
	\bibitem[{Varia et~al.(2023)Varia, Wang, Halder, Vacareanu, Ballesteros,
		Benajiba, John, Anubhai, Muresan \& Roth}]{ITMIT}
	\bibinfo{author}{Varia, S.}, \bibinfo{author}{Wang, S.},
	\bibinfo{author}{Halder, K.}, \bibinfo{author}{Vacareanu, R.},
	\bibinfo{author}{Ballesteros, M.}, \bibinfo{author}{Benajiba, Y.},
	\bibinfo{author}{John, N.~A.}, \bibinfo{author}{Anubhai, R.},
	\bibinfo{author}{Muresan, S.}, \& \bibinfo{author}{Roth,
		D.}(\textcharter{\bibinfo{year}{2023}}).
	\newblock \bibinfo{title}{Instruction tuning for few-shot aspect-based
		sentiment analysis}.
	\newblock In {\it \bibinfo{booktitle}{Proceedings of the 13th Workshop on
			Computational Approaches to Subjectivity, Sentiment, {\&} Social Media
			Analysis, WASSA{@}ACL, July 14, 2023}\/} (pp. \bibinfo{pages}{19--27}).
	\newblock \bibinfo{address}{Toronto, Canada}: \bibinfo{publisher}{ACL}.
	\newblock \DOIprefix\doi{10.18653/V1/2023.WASSA-1.3}.
	\bibitem[{Vaswani et~al.(2017)Vaswani, Shazeer, Parmar, Uszkoreit, Jones,
		Gomez, Kaiser \& Polosukhin}]{transformer}
	\bibinfo{author}{Vaswani, A.}, \bibinfo{author}{Shazeer, N.},
	\bibinfo{author}{Parmar, N.}, \bibinfo{author}{Uszkoreit, J.},
	\bibinfo{author}{Jones, L.}, \bibinfo{author}{Gomez, A.~N.},
	\bibinfo{author}{Kaiser, L.}, \& \bibinfo{author}{Polosukhin,
		I.}(\textcharter{\bibinfo{year}{2017}}).
	\newblock \bibinfo{title}{Attention is all you need}.
	\newblock In {\it \bibinfo{booktitle}{Advances in Neural Information Processing
			Systems 30: Annual Conference on Neural Information Processing Systems, NIPS,
			December 4-9, 2017}\/} (pp. \bibinfo{pages}{5998--6008}).
	\newblock \bibinfo{address}{Long Beach, CA, USA}: \bibinfo{publisher}{ACM}.
	\bibitem[{Wang et~al.(2024)Wang, Xu, Ding, Liang \& Xu}]{202401}
	\bibinfo{author}{Wang, Q.}, \bibinfo{author}{Xu, H.}, \bibinfo{author}{Ding,
		K.}, \bibinfo{author}{Liang, B.}, \& \bibinfo{author}{Xu,
		R.}(\textcharter{\bibinfo{year}{2024}}).
	\newblock \bibinfo{title}{In-context example retrieval from multi-perspectives
		for few-shot aspect-based sentiment analysis}.
	\newblock In {\it \bibinfo{booktitle}{Proceedings of the 2024 Joint
			International Conference on Computational Linguistics, Language Resources and
			Evaluation, {LREC/COLING}}\/} (pp. \bibinfo{pages}{8975--8985}).
	\newblock \bibinfo{address}{Torino, Italy}: \bibinfo{publisher}{{ELRA} and
		{ICCL}}.
	\bibitem[{Wu et~al.(2020{\natexlab{a}})Wu, Ying, Zhao, Fan, Dai \& Xia}]{gts}
	\bibinfo{author}{Wu, Z.}, \bibinfo{author}{Ying, C.}, \bibinfo{author}{Zhao,
		F.}, \bibinfo{author}{Fan, Z.}, \bibinfo{author}{Dai, X.}, \&
	\bibinfo{author}{Xia, R.}(\textcharter{\bibinfo{year}{2020}{\natexlab{a}}}).
	\newblock \bibinfo{title}{Grid tagging scheme for aspect-oriented fine-grained
		opinion extraction}.
	\newblock In {\it \bibinfo{booktitle}{Findings of the Association for
			Computational Linguistics, {EMNLP}, November, 2020}\/} (pp.
	\bibinfo{pages}{2576--2585}).
	\newblock \bibinfo{address}{Online}: \bibinfo{publisher}{ACL}.
	\newblock \DOIprefix\doi{10.18653/v1/2020.findings-emnlp.234}.
	\bibitem[{Wu et~al.(2020{\natexlab{b}})Wu, Zhao, Dai, Huang \& Chen}]{AOPE_2}
	\bibinfo{author}{Wu, Z.}, \bibinfo{author}{Zhao, F.}, \bibinfo{author}{Dai,
		X.}, \bibinfo{author}{Huang, S.}, \& \bibinfo{author}{Chen,
		J.}(\textcharter{\bibinfo{year}{2020}{\natexlab{b}}}).
	\newblock \bibinfo{title}{Latent opinions transfer network for target-oriented
		opinion words extraction}.
	\newblock In {\it \bibinfo{booktitle}{The Thirty-Fourth {AAAI} Conference on
			Artificial Intelligence, {AAAI}, The Thirty-Second Innovative Applications of
			Artificial Intelligence Conference, {IAAI}, The Tenth {AAAI} Symposium on
			Educational Advances in Artificial Intelligence, {EAAI}, February 7-12,
			2020}\/} (pp. \bibinfo{pages}{9298--9305}).
	\newblock \bibinfo{address}{New York, NY, USA}: \bibinfo{publisher}{{AAAI}
		Press}.
	\newblock \DOIprefix\doi{10.1609/AAAI.V34I05.6469}.
	\bibitem[{Xu et~al.(2021)Xu, Chia \& Bing}]{span-aste}
	\bibinfo{author}{Xu, L.}, \bibinfo{author}{Chia, Y.~K.}, \&
	\bibinfo{author}{Bing, L.}(\textcharter{\bibinfo{year}{2021}}).
	\newblock \bibinfo{title}{Learning span-level interactions for aspect sentiment
		triplet extraction}.
	\newblock In {\it \bibinfo{booktitle}{Proceedings of the 59th Annual Meeting of
			the Association for Computational Linguistics (ACL), The 11th International
			Joint Conference on Natural Language Processing (IJCNLP)}\/} (pp.
	\bibinfo{pages}{4755--4766}).
	\newblock \bibinfo{address}{Online}: \bibinfo{publisher}{ACL}.
	\newblock \DOIprefix\doi{10.18653/v1/2021.acl-long.367}.
	\bibitem[{Xu et~al.(2020)Xu, Li, Lu \& Bing}]{jet}
	\bibinfo{author}{Xu, L.}, \bibinfo{author}{Li, H.}, \bibinfo{author}{Lu, W.},
	\& \bibinfo{author}{Bing, L.}(\textcharter{\bibinfo{year}{2020}}).
	\newblock \bibinfo{title}{Position-aware tagging for aspect sentiment triplet
		extraction}.
	\newblock In {\it \bibinfo{booktitle}{Proceedings of the 2020 Conference on
			Empirical Methods in Natural Language Processing, {EMNLP}, November 16-20,
			2020}\/} (pp. \bibinfo{pages}{2339--2349}).
	\newblock \bibinfo{address}{Online}: \bibinfo{publisher}{ACL}.
	\newblock \DOIprefix\doi{10.18653/v1/2020.emnlp-main.183}.
	\bibitem[{Yang et~al.(2022)Yang, Gan, Wang, Hu, Lu, Liu \& Wang}]{kbvqa1}
	\bibinfo{author}{Yang, Z.}, \bibinfo{author}{Gan, Z.}, \bibinfo{author}{Wang,
		J.}, \bibinfo{author}{Hu, X.}, \bibinfo{author}{Lu, Y.},
	\bibinfo{author}{Liu, Z.}, \& \bibinfo{author}{Wang,
		L.}(\textcharter{\bibinfo{year}{2022}}).
	\newblock \bibinfo{title}{An empirical study of {GPT-3} for few-shot
		knowledge-based {VQA}}.
	\newblock In {\it \bibinfo{booktitle}{Thirty-Sixth {AAAI} Conference on
			Artificial Intelligence, {AAAI} 2022, Thirty-Fourth Conference on Innovative
			Applications of Artificial Intelligence, {IAAI} 2022, The Twelveth Symposium
			on Educational Advances in Artificial Intelligence, {EAAI} 2022, February 22
			- March 1, 2022}\/} (pp. \bibinfo{pages}{3081--3089}).
	\newblock \bibinfo{address}{Online}: \bibinfo{publisher}{{AAAI} Press}.
	\newblock \DOIprefix\doi{10.1609/AAAI.V36I3.20215}.
	\bibitem[{Yin et~al.(2016)Yin, Wei, Dong, Xu, Zhang \& Zhou}]{ae2}
	\bibinfo{author}{Yin, Y.}, \bibinfo{author}{Wei, F.}, \bibinfo{author}{Dong,
		L.}, \bibinfo{author}{Xu, K.}, \bibinfo{author}{Zhang, M.}, \&
	\bibinfo{author}{Zhou, M.}(\textcharter{\bibinfo{year}{2016}}).
	\newblock \bibinfo{title}{Unsupervised word and dependency path embeddings for
		aspect term extraction}.
	\newblock In {\it \bibinfo{booktitle}{Proceedings of the Twenty-Fifth
			International Joint Conference on Artificial Intelligence, {IJCAI}, 9-15 July
			2016}\/} (pp. \bibinfo{pages}{2979--2985}).
	\newblock \bibinfo{address}{New York, NY, USA}:
	\bibinfo{publisher}{{IJCAI/AAAI} Press}.
	\bibitem[{Zhang et~al.(2019)Zhang, Li \& Song}]{ALSC_3}
	\bibinfo{author}{Zhang, C.}, \bibinfo{author}{Li, Q.}, \&
	\bibinfo{author}{Song, D.}(\textcharter{\bibinfo{year}{2019}}).
	\newblock \bibinfo{title}{Aspect-based sentiment classification with
		aspect-specific graph convolutional networks}.
	\newblock In {\it \bibinfo{booktitle}{Proceedings of the 2019 Conference on
			Empirical Methods in Natural Language Processing, {EMNLP}, The 9th
			International Joint Conference on Natural Language Processing, {IJCNLP},
			November 3-7, 2019}\/} (pp. \bibinfo{pages}{4567--4577}).
	\newblock \bibinfo{address}{Hong Kong, China}: \bibinfo{publisher}{ACL}.
	\newblock \DOIprefix\doi{10.18653/V1/D19-1464}.
	\bibitem[{Zhang et~al.(2023{\natexlab{a}})Zhang, Li, Deng, Bing \&
		Lam}]{absasurvey}
	\bibinfo{author}{Zhang, W.}, \bibinfo{author}{Li, X.}, \bibinfo{author}{Deng,
		Y.}, \bibinfo{author}{Bing, L.}, \& \bibinfo{author}{Lam,
		W.}(\textcharter{\bibinfo{year}{2023}{\natexlab{a}}}).
	\newblock \bibinfo{title}{A survey on aspect-based sentiment analysis: Tasks,
		methods, and challenges}.
	\newblock {\it \bibinfo{journal}{IEEE Transactions on Knowledge and Data
			Engineering}\/},  {\it \bibinfo{volume}{35}\/},
	\bibinfo{pages}{11019--11038}. \DOIprefix\doi{10.1109/TKDE.2022.3230975}.
	\bibitem[{Zhang et~al.(2023{\natexlab{b}})Zhang, Xu, Cai, Tan \& Zhu}]{alsc3}
	\bibinfo{author}{Zhang, X.}, \bibinfo{author}{Xu, J.}, \bibinfo{author}{Cai,
		Y.}, \bibinfo{author}{Tan, X.}, \& \bibinfo{author}{Zhu,
		C.}(\textcharter{\bibinfo{year}{2023}{\natexlab{b}}}).
	\newblock \bibinfo{title}{Detecting dependency-related sentiment features for
		aspect-level sentiment classification}.
	\newblock {\it \bibinfo{journal}{IEEE Transactions on Affective Computing}\/},
	{\it \bibinfo{volume}{14}\/}, \bibinfo{pages}{196--210}.
	\DOIprefix\doi{10.1109/TAFFC.2021.3063259}.
	\bibitem[{Zhang et~al.(2022)Zhang, Peng, Han, Han, Yue \& Liu}]{aoe3}
	\bibinfo{author}{Zhang, Y.}, \bibinfo{author}{Peng, T.}, \bibinfo{author}{Han,
		R.}, \bibinfo{author}{Han, J.}, \bibinfo{author}{Yue, L.}, \&
	\bibinfo{author}{Liu, L.}(\textcharter{\bibinfo{year}{2022}}).
	\newblock \bibinfo{title}{Synchronously tracking entities and relations in a
		syntax-aware parallel architecture for aspect-opinion pair extraction}.
	\newblock {\it \bibinfo{journal}{Applied Intelligence}\/},  {\it
		\bibinfo{volume}{52}\/}, \bibinfo{pages}{15210--15225}.
	\DOIprefix\doi{10.1007/s10489-022-03286-w}.
	\bibitem[{Zhao et~al.(2023)Zhao, Jin, Ser \& Yang}]{chatagri}
	\bibinfo{author}{Zhao, B.}, \bibinfo{author}{Jin, W.}, \bibinfo{author}{Ser,
		J.~D.}, \& \bibinfo{author}{Yang, G.}(\textcharter{\bibinfo{year}{2023}}).
	\newblock \bibinfo{title}{Chatagri: Exploring potentials of chatgpt on
		cross-linguistic agricultural text classification}.
	\newblock {\it \bibinfo{journal}{Neurocomputing}\/},  {\it
		\bibinfo{volume}{557}\/}, \bibinfo{pages}{126708}.
	\DOIprefix\doi{10.1016/J.NEUCOM.2023.126708}.
	\bibitem[{Zhou et~al.(2023)Zhou, Li, Li, Yu, Liu, Wang, Zhang, Ji, Yan, He,
		Peng, Li, Wu, Liu, Xie, Xiong, Pei, Yu \& Sun}]{now2}
	\bibinfo{author}{Zhou, C.}, \bibinfo{author}{Li, Q.}, \bibinfo{author}{Li, C.},
	\bibinfo{author}{Yu, J.}, \bibinfo{author}{Liu, Y.}, \bibinfo{author}{Wang,
		G.}, \bibinfo{author}{Zhang, K.}, \bibinfo{author}{Ji, C.},
	\bibinfo{author}{Yan, Q.}, \bibinfo{author}{He, L.}, \bibinfo{author}{Peng,
		H.}, \bibinfo{author}{Li, J.}, \bibinfo{author}{Wu, J.},
	\bibinfo{author}{Liu, Z.}, \bibinfo{author}{Xie, P.}, \bibinfo{author}{Xiong,
		C.}, \bibinfo{author}{Pei, J.}, \bibinfo{author}{Yu, P.~S.}, \&
	\bibinfo{author}{Sun, L.}(\textcharter{\bibinfo{year}{2023}}).
	\newblock \bibinfo{title}{A comprehensive survey on pretrained foundation
		models: A history from bert to chatgpt}.
	\newblock \href{http://arxiv.org/abs/2302.09419}{\tt arXiv:2302.09419}.
	
\end{thebibliography}

\end{document}